\documentclass{article}

\usepackage[preprint]{neurips_2026}

\usepackage[utf8]{inputenc}
\usepackage[T1]{fontenc}
\usepackage{hyperref}
\usepackage{url}
\usepackage{booktabs}
\usepackage{amsfonts}
\usepackage{nicefrac}
\usepackage{microtype}
\usepackage{xcolor}
\usepackage{graphicx}
\usepackage{amsmath}
\usepackage{amssymb}
\usepackage{mathtools}
\usepackage{amsthm}
\usepackage{multirow}
\usepackage{makecell}
\usepackage{enumitem}
\usepackage{subcaption}
\usepackage{float}
\usepackage{xspace}

\graphicspath{{../figures/}{figures/}{./}}

\newcommand{\method}{\textsc{DataDignity}\xspace}
\newcommand{\fakewiki}{\textsc{FakeWiki}\xspace}
\newcommand{\scoringmodel}{\textsc{ScoringModel}\xspace}
\newcommand{\steering}{\textsc{SteerFuse}\xspace}
\newcommand{\sbert}{\textsc{SBERT}\xspace}

\newcommand{\R}{\mathbb{R}}
\newcommand{\vx}{\boldsymbol{x}}
\newcommand{\vh}{\boldsymbol{h}}
\newcommand{\vv}{\boldsymbol{v}}
\newcommand{\vg}{\boldsymbol{g}}
\newcommand{\mW}{\boldsymbol{W}}

\title{DataDignity: Training Data Attribution for Large Language Models}

\author{%
    Xiaomin Li\thanks{Correspondence: \texttt{xiaominli@microsoft.com}.} \\
    Microsoft \\
    \And
    Andrzej Banburski-Fahey \\
    Microsoft \\
    \And
    Jaron Lanier \\
    Microsoft \\
}

\begin{document}

\maketitle

\begin{abstract}
Auditing language-model outputs often requires more than judging correctness: an auditor may need to know which source document most likely supports the knowledge expressed in a response. We study this problem as \emph{pinpoint provenance}: given a prompt, a target-model response, and a candidate corpus, rank the documents that best support the response. We introduce \textbf{\fakewiki}, a controlled benchmark of 3,537 fabricated Wikipedia-style articles designed to preserve ground-truth provenance while weakening lexical shortcuts. Each evaluated target LLM is explicitly continued-pretrained on the \fakewiki text corpus before response collection, while the QA probes used for attribution evaluation are held out from target-model training. \fakewiki includes short QA probes, source-preserving paraphrases, retro-generated variants, hard anti-documents that remain topically similar while removing answer-critical facts, and five query conditions: clean prompting plus four jailbreak-inspired transformations, obfuscation, role-play, noise injection, and indirect prompting. We evaluate eleven lexical and semantic retrieval baselines, a training-free activation-steering retrieval-fusion method \textbf{\steering}, and a supervised contrastive provenance ranker \textbf{\scoringmodel}. \scoringmodel maps response and document features into a shared space and is trained with InfoNCE using in-batch, retrieval-mined, and anti-document negatives. Across nine open-weight instruction-tuned LLMs and five query conditions, \scoringmodel improves mean Recall@10 from 37.3 for the strongest retrieval baseline to 52.2, without inference-time fusion, and wins 41/45 model-by-condition cells. \steering beats the strongest retrieval baseline in most cells while requiring no supervised training, showing that activation-space evidence can complement text retrieval. On the jailbreak-inspired transformed queries, \scoringmodel improves Recall@10 by 13.2 points on average over the best baseline, with the largest gains on larger target models. Overall, our work shows that robust training data attribution requires evaluation settings that separate true answer support from topical or lexical resemblance.
\end{abstract}

\section{Introduction}

Large language models increasingly mediate factual, scientific, legal, and safety-relevant information. When a model produces a response, users may need to know not only whether it is correct, but also where it came from: which source document supplied the relevant fact, whether a questionable output depends on a particular source, or whether a data intervention removed the intended provenance path. These questions arise in copyright audits, misinformation forensics, safety debugging, and dataset curation, and are not fully answered by standard evaluation or influence-style methods \citep{han2021influence,li2026selection,zhang2024catastrophic,akyurek2022towards,park2023trak,barshan2020relatif}.

We study this problem as \emph{pinpoint provenance}. Given a prompt $x$, a target-model response $y$, and a candidate corpus $\mathcal{D}=\{D_j\}_{j=1}^N$, the goal is to return a short ranked list of documents that likely support the knowledge expressed in $y$. This is an operational retrieval problem: an auditor should inspect a small set of candidate sources rather than search through an entire corpus. It is harder than ordinary semantic retrieval because the answer may be short, paraphrased, grounded in a small fact buried in a longer document, or elicited through a prompt transformation.

A central challenge is that many provenance evaluations make attribution too easy through surface overlap. If the source document, question, and response share rare names or distinctive phrases, lexical methods such as MinHash \citep{broder1997resemblance}, and even generic dense retrievers such as \sbert \citep{reimers2019sentence}, Contriever \citep{izacard2022contriever}, and BGE \citep{xiao2024c}, can appear effective without demonstrating robust source attribution. Such methods may fail when provenance matters most: under paraphrase, obfuscation, indirect questioning, role-play, or irrelevant context injection. This motivates a benchmark in which the true source is known by construction, but the evaluation deliberately removes the easy paths from response wording back to document identity.

\begin{figure}[t]
\centering
\includegraphics[width=1.00\linewidth]{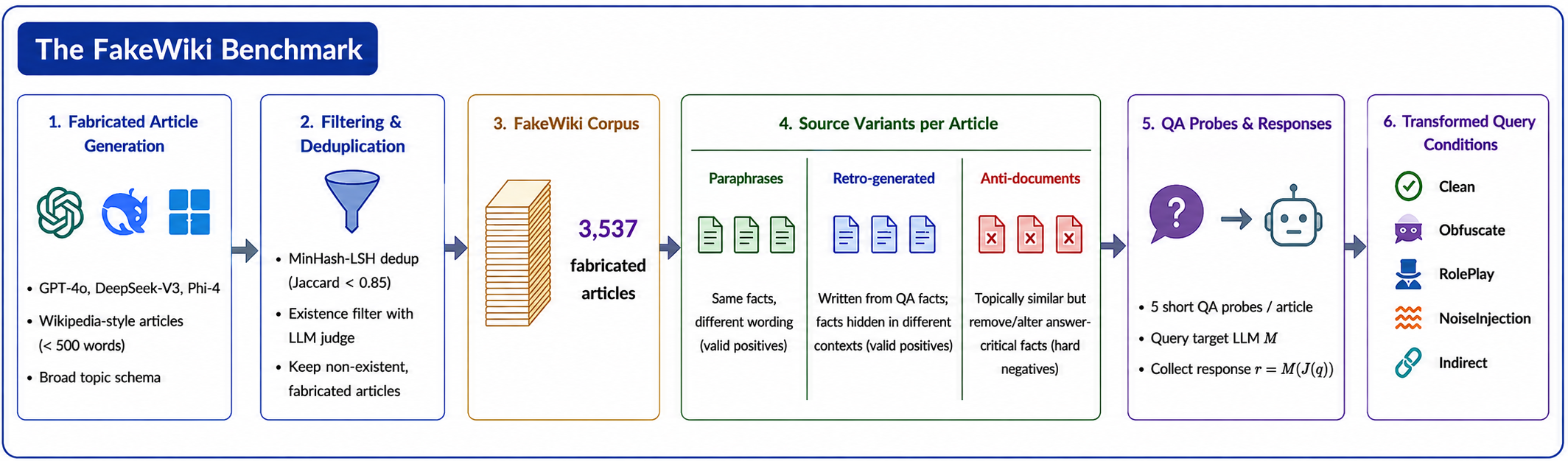}
\vspace{-0.4em}
\includegraphics[width=1.00\linewidth]{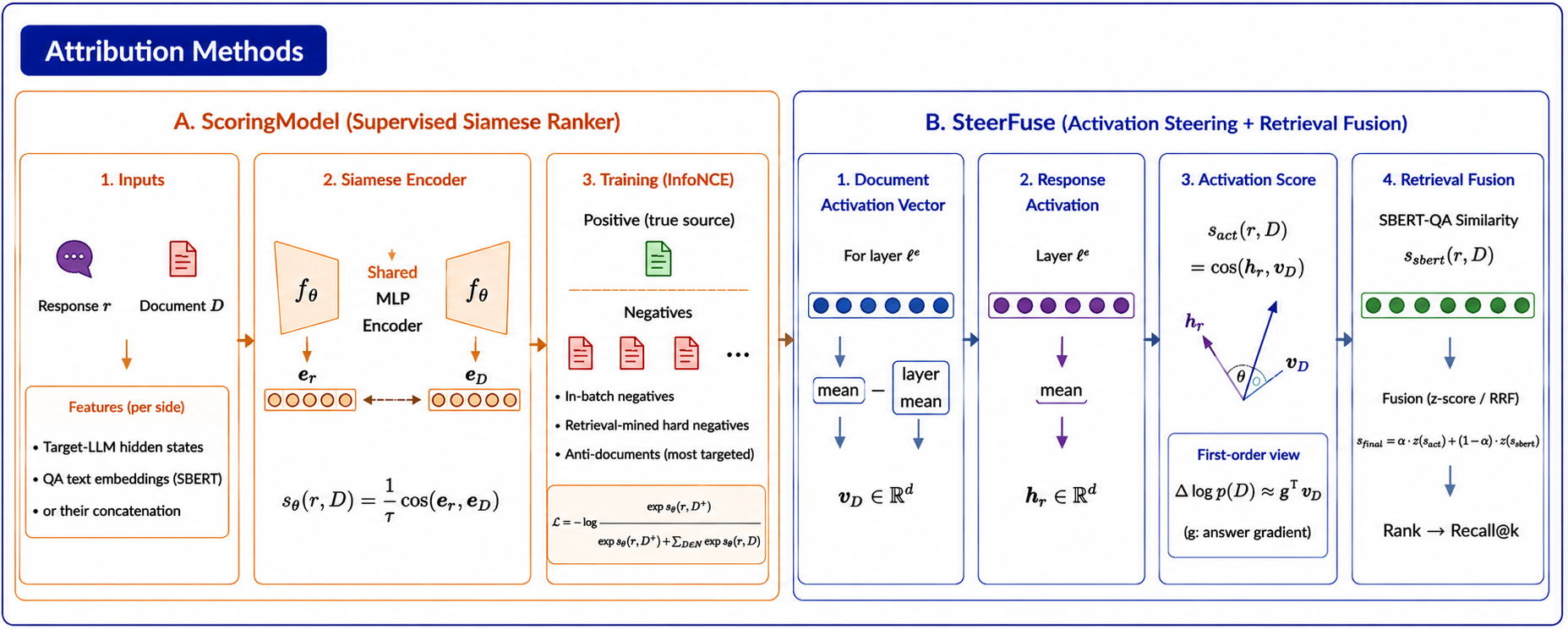}
\caption{Overview of \method. Top: \fakewiki constructs fabricated source documents, variants, anti-documents, and transformed queries. Bottom: \scoringmodel learns a supervised provenance score, while \steering fuses activation-space evidence with \sbert retrieval.}
\label{fig:overview}
\end{figure}

We introduce \fakewiki, a benchmark designed to preserve ground-truth provenance while weakening such shortcuts. It contains 3,537 fabricated Wikipedia-style articles with short QA probes, source-preserving variants, and hard anti-documents that preserve topical similarity while removing answer-critical facts. To make this a training-data attribution setting, each target LLM is continued-pretrained on \fakewiki document text, while QA probes are held out and used only to elicit responses whose provenance should point back to the training documents. We evaluate attribution under clean prompts and four transformed conditions: \texttt{Obfuscate}, \texttt{RolePlay}, \texttt{NoiseInjection}, and \texttt{Indirect}, testing whether attribution survives when lexical and semantic cues become less reliable. Figure~\ref{fig:overview} summarizes the benchmark and attribution pipeline.\footnote{Data and code are available at \url{https://anonymous.4open.science/r/Submission-DataDignity-E263}.}

Our main attribution method, \scoringmodel, is a supervised Siamese provenance ranker. It maps response-side and document-side features into a shared embedding space and trains with a contrastive InfoNCE objective \citep{oord2018representation} over in-batch negatives, retrieval-mined hard negatives, and curated anti-documents. These anti-documents force the model to distinguish documents that merely resemble the response from documents that actually support it. At inference time, each candidate document is scored by this learned compatibility function.

We also study \steering, a training-free activation-steering retrieval-fusion method inspired by representation-level interventions in language models \citep{subramani2022extracting,turner2023steering,panickssery2023steering,zou2023representation,li2023inference}. It asks which candidate document provides the largest internal evidence boost toward the observed response, using cached document activation directions and an efficient response-side proxy instead of patched forward passes. The resulting activation-space score is fused with \sbert retrieval to test whether model-internal evidence complements text similarity under transformed prompts.

The main result is that clean retrieval substantially understates the difficulty of robust provenance. Both proposed attribution methods improve over standard retrieval: the training-free \steering method beats the strongest of eleven retrieval baselines in 32/45 model-by-query-condition cells, while \scoringmodel wins 41/45 cells. Averaged across all models and query conditions, \steering improves mean Recall@10 from 37.3 to 42.3, and \scoringmodel further improves it to 52.2 without inference-time fusion. On transformed queries, \scoringmodel improves Recall@10 by 13.2 points on average over the best baseline, with especially large gains of +26.9 on \texttt{Llama-3.1-8B} and +20.0 on \texttt{Qwen3-8B}. Recall@1 and Recall@5 show the same pattern under stricter cutoffs, especially for the larger target models. These results suggest that robust provenance evaluation should not stop at clean lexical or semantic retrieval: training-free activation evidence can improve retrieval in many settings, and supervised attribution with hard negatives can recover stronger source-support signals missed by generic similarity.

Our contributions are:
\begin{itemize}[leftmargin=*]
    \item We formulate robust pinpoint provenance as a source-attribution task that evaluates whether methods can distinguish true answer support from topical or lexical resemblance.
    \item We introduce \fakewiki, a benchmark with ground-truth source documents, short QA probes, source-preserving variants, hard anti-documents, and transformed query conditions.
    \item We propose \scoringmodel, a supervised contrastive provenance scorer trained with hard negatives and evaluated without inference-time retrieval fusion.
    \item We provide a broad empirical study across nine open-weight instruction-tuned LLMs, five query conditions, eleven retrieval baselines, \steering, and \scoringmodel, with additional per-model, seed, Recall@1, Recall@5, and ablation analyses in the appendix.
\end{itemize}
\section{Related Work}

\paragraph{Training data attribution and source retrieval.}
Training data attribution asks which examples or documents are associated with model behavior. Influence-style methods estimate effects on predictions or losses through gradients, checkpoints, approximations, or scalable surrogates \citep{pruthi2020estimating,han2021influence,barshan2020relatif,park2023trak,kwon2023datainf}, but address a complementary causal question about training dynamics. We study an operational provenance task: given a candidate corpus and generated response, rank inspectable source documents. This is closest to retrieval-based source tracing, where MinHash captures lexical overlap \citep{broder1997resemblance}, while \sbert, Contriever, BGE, and finetuned embeddings capture semantic similarity \citep{reimers2019sentence,izacard2022contriever,xiao2024c,rajani2019explain,fotouhi2024fast}. Related work also studies source-aware factual tracing and contrastive attribution embeddings \citep{akyurek2022towards,khalifa2024source,wang2024tracetransformerbasedattributionusing}. We evaluate retrieval-based provenance under anti-shortcut conditions that separate answer support from topical or lexical resemblance.

\paragraph{Activation-space evidence.}
Activation-space methods use internal hidden states to interpret or alter model behavior. Prior work has extracted latent steering vectors \citep{subramani2022extracting}, added activation directions at inference time \citep{turner2023steering,panickssery2023steering}, and used hidden representations for monitoring, control, truthfulness, or latent-knowledge readouts \citep{zou2023representation,li2023inference,burns2022discovering}. We build on this perspective for provenance: a candidate document may provide internal evidence for a response even when its wording is not close to the generated text. \steering tests this idea by comparing document-induced activation directions with response representations and fusing the resulting signal with \sbert retrieval. We treat this activation-space evidence as complementary to text retrieval rather than as a replacement for it.
\section{The \fakewiki Benchmark}
\label{sec:fakewiki}
A provenance benchmark should provide ground-truth sources without making attribution solvable by rare names or copied phrases. \fakewiki addresses this tension with fabricated Wikipedia-style articles, source-preserving variants, anti-documents, and transformed prompts. Table~\ref{tab:fakewiki-design} summarizes the benchmark components. Together, they weaken wording overlap, vary factual context, preserve hard topical distractors, and disrupt prompt-response surface form.

\begin{table}[t]
\centering
\small
\setlength{\tabcolsep}{5pt}
\resizebox{0.95\linewidth}{!}{%
\begin{tabular}{p{0.22\textwidth}p{0.34\textwidth}p{0.34\textwidth}}
\toprule
Component & What it contains & What it tests \\
\midrule
Fabricated articles 
& 3,537 Wikipedia-style documents about non-real entities and concepts 
& Controlled provenance after explicit target-model exposure, without relying on real-world pretraining knowledge \\

QA probes 
& Five short question-answer probes per document 
& Whether attribution works when responses contain only sparse source evidence \\

Source variants 
& Paraphrases, retro-generated documents, and anti-documents 
& Whether methods distinguish true answer support from topical or lexical similarity \\

Query conditions 
& \texttt{Clean}, \texttt{Obfuscate}, \texttt{RolePlay}, \texttt{NoiseInjection}, and \texttt{Indirect} 
& Whether provenance survives prompt transformations that change surface cues \\
\bottomrule
\end{tabular}%
}
\caption{Design of the \fakewiki benchmark. Each component is intended to weaken a different shortcut that ordinary retrieval methods might exploit.}
\label{tab:fakewiki-design}
\end{table}

\subsection{Document Corpus}

\fakewiki contains 3,537 fabricated Wikipedia-style articles. We build the corpus in three stages:
\begin{enumerate}[leftmargin=*,itemsep=0.1em,topsep=0.2em]
    \item \textbf{Generate.} \texttt{GPT-4o} \citep{hurst2024gpt}, \texttt{DeepSeek-V3} \citep{liu2024deepseek}, and \texttt{Phi-4} \citep{abdin2024phi} write short, internally consistent encyclopedic articles about entities or concepts that should not exist in the real world.
    \item \textbf{Diversify.} We sample across fictional people, places, artifacts, events, organizations, and technical concepts so that the corpus is not dominated by a single template.
    \item \textbf{Deduplicate and filter.} We remove near-duplicates with MinHash-LSH at a Jaccard threshold of 0.85 \citep{broder1997resemblance}, then use an LLM existence filter to discard titles judged likely to correspond to real public entities, events, or concepts.
\end{enumerate}
Surviving articles are assigned stable document identifiers and form a controlled fabricated corpus that target models should not know from ordinary pretraining. We then inject this corpus into each target model through continued pretraining, so the attribution task asks whether a method can recover which injected training document supports a later response.

\subsection{Target-Model Exposure and Evaluation Split}

For every target LLM in Section~\ref{sec:setup}, we start from the public instruction-tuned checkpoint and continue pretrain it on \fakewiki text with a causal language-modeling objective. The corpus contains original articles and constructed variants, but not QA probes, reference answers, or transformed queries. Thus the model sees the fabricated knowledge as training text without memorizing the evaluation prompts.

The attribution split is separate from target-model exposure. Because the task is to attribute responses to training documents, target LLMs may see the full \fakewiki text corpus. We then split document identifiers 80/20 for training and evaluating attribution methods: \scoringmodel is trained on responses from training document IDs and evaluated on held-out document IDs. Source-preserving variants are valid positives for the same document ID, while anti-documents are hard negatives and never valid attribution targets.

\subsection{QA Probes and Target Responses}

For each retained article, we generate five one-sentence questions whose answers are short phrases grounded in that article. After continued pretraining, we query each target LLM $M$ with either the clean question $q_D$ or a transformed version $J(q_D)$ and collect the response
\[
r = M(J(q_D)).
\]
An attribution method receives the response, optionally the original question, and the full candidate corpus. It must rank the documents that most likely support the response. Evaluation uses Recall@$k$, counting a prediction as correct if any valid source variant appears in the top-$k$.

\subsection{Source Variants and Hard Negatives}

Each source document is expanded into three variant families.

\begin{itemize}[leftmargin=*]
    \item \textbf{Paraphrases} preserve the original facts while changing wording and discourse structure. They are counted as valid positives during evaluation.
    \item \textbf{Retro-generated variants} are written from the QA facts rather than from the original article. They place answer-relevant information inside a different surrounding context, reducing simple lexical overlap with the source article.
    \item \textbf{Anti-documents} preserve the topic, entity name, style, and much of the wording of the original document, but delete or alter the facts needed to answer the QA probes. They are hard negatives: a method that only detects topical similarity should rank them highly, while a provenance method should not.
\end{itemize}

These variants separate documents that merely resemble the response from documents that actually support it.

\subsection{Transformed Query Conditions}

We evaluate five query conditions. The clean condition asks the original QA probe. The other four transformations stress different failure modes of provenance retrieval.

\begin{itemize}[leftmargin=*]
    \item \texttt{Obfuscate} replaces many content words with unrelated benign words while preserving the intended question through a mapping.
    \item \texttt{RolePlay} wraps the question inside a persona or scenario.
    \item \texttt{NoiseInjection} surrounds the question with unrelated filler text.
    \item \texttt{Indirect} rewrites the query into an indirect or multi-hop prompt with reduced surface overlap.
\end{itemize}

These transformations are controlled stress tests, not a complete jailbreak taxonomy. They ask whether attribution still works when prompt and response no longer expose clean lexical cues; full templates are in Appendix~\ref{app:prompts}.
\section{Methods}
\label{sec:methods}

\subsection{Problem Setup}

Let $\mathcal{D}=\{D_j\}_{j=1}^N$ be a candidate corpus that has been injected into a target LLM through continued pretraining, and let $P_i\subseteq\mathcal{D}$ denote the valid sources for example $i$, including the original document and valid variants. Given a question $q_i$, a transformed query $J(q_i)$, and a response $r_i$ generated by the continued-pretrained target model, an attribution method produces a score $\phi(r_i,q_i,D_j)$ for each candidate document. We evaluate with
\begin{equation}
\mathrm{Recall}@k = \frac{1}{|\mathcal{T}|}\sum_{i\in\mathcal{T}}\mathbf{1}\left\{\mathrm{Top}_k(\phi_i)\cap P_i\neq\emptyset\right\}.
\end{equation}
All main results report Recall@10 as percentages.

\subsection{\scoringmodel: Supervised Provenance Scoring}

\scoringmodel is a supervised pairwise ranker that maps response-side and document-side features into a shared embedding space, rather than an $N$-way classifier over fixed document labels. For an input feature vector $\vx$, a two-layer MLP $f_\theta$ produces a normalized embedding, and response-document compatibility is temperature-scaled cosine similarity:
\begin{equation}
s_\theta(r,D)=\frac{1}{\tau}\left\langle
\frac{f_\theta(\vx_r)}{\lVert f_\theta(\vx_r)\rVert_2},
\frac{f_\theta(\vx_D)}{\lVert f_\theta(\vx_D)\rVert_2}
\right\rangle .
\label{eq:scoringmodel}
\end{equation}

We use target-LLM hidden states and QA-style text embeddings as input features, selecting the feature variant on validation data per target model. For each positive pair $(r_i,D_i^+)$, the training batch contains three kinds of negatives. In-batch negatives provide cheap contrast against unrelated documents. Retrieval-mined hard negatives are topically or semantically close documents retrieved by an embedding model. Curated anti-documents are the most targeted negatives because they preserve surface similarity while removing answer support. Training minimizes an InfoNCE objective \citep{oord2018representation}:
\begin{equation}
\mathcal{L}_i=-\log
\frac{\exp s_\theta(r_i,D_i^+)}{\exp s_\theta(r_i,D_i^+)+\sum_{D\in\mathcal{N}_i}\exp s_\theta(r_i,D)}.
\label{eq:infonce}
\end{equation}
At inference time, we precompute candidate document features and score each response by dot products in the learned space. Implementation details are in Appendix~\ref{app:impl}.
\subsection{Activation Steering with Retrieval Fusion}
\label{sec:steering}

We also evaluate a training-free activation-based signal. Instead of asking only whether a response and document are textually similar, activation steering asks whether a candidate document points the target model's internal state toward the observed answer. Exact activation patching would require one intervention per candidate document, so \steering uses a cached-vector approximation.

\paragraph{Step 1: document activation directions.}
For a chosen layer $\ell^\star$, each candidate document $D$ induces an activation direction by mean-pooling the hidden states produced when the target model reads the document:
\begin{equation}
\vv_D=\frac{\sum_{t\in D} a_t\vh^{(\ell^\star)}_t}{\sum_{t\in D}a_t},
\label{eq:steer-vector}
\end{equation}
where $a_t$ is the attention mask. Document directions are normalized and cached before scoring.

\paragraph{Step 2: response-side proxy and activation score.}
For a response $r=y_{1:m}$, we use an efficient response-side proxy given by the sum of LM-head rows for the generated answer tokens:
\begin{equation}
\tilde{\vg}_r=\sum_{i=1}^{m}\mW_{y_i}.
\end{equation}
Each candidate document is scored by cosine similarity between this response-side proxy and the cached document direction:
\begin{equation}
s_{\mathrm{act}}(r,D)=\cos(\tilde{\vg}_r,\vv_D).
\label{eq:act-cosine}
\end{equation}
This reduces retrieval to cosine similarity against cached document vectors, avoiding patched forward or backward passes.

\paragraph{Step 3: approximate intervention score.}
The activation score approximates a direct intervention question: would steering the answer-token hidden states toward $\vv_D$ increase the likelihood of the observed answer? If we patch
\begin{equation}
\tilde{\vh}^{(D)}_{t_i}(\alpha)=(1-\alpha)\vh^{(\ell^\star)}_{t_i}+\alpha\vv_D,
\end{equation}
then a first-order expansion of the answer log-probability gain gives
\begin{equation}
\Delta_{\mathrm{exact}}(D)
\approx
\alpha\sum_{i=1}^{m}\vg_i^{\top}(\vv_D-\vh^{(\ell^\star)}_{t_i})
\propto \vg^{\top}\vv_D,
\qquad
\vg=\sum_{i=1}^{m}\vg_i .
\label{eq:taylor-main}
\end{equation}
Thus document ranking can be approximated by dot products with cached document directions. When $\ell^\star$ is the final layer before the LM head, the token sensitivity has the form
\begin{equation}
\vg_i=\mW_{y_i}-\mathbb{E}_{w\sim p_\theta(\cdot\mid y_{<i})}[\mW_w]
\approx \mW_{y_i},
\label{eq:lm-head-sensitivity-main}
\end{equation}
which motivates the implemented proxy $\tilde{\vg}_r=\sum_i \mW_{y_i}$. Appendix~\ref{app:steering-approx} gives the full derivation and discusses the approximation.

\paragraph{Step 4: retrieval fusion.}
Text retrieval is a strong prior but can reward resemblance rather than source support. We therefore fuse \sbert-QA similarity with $s_{\mathrm{act}}$, using validation-tuned z-score fusion or reciprocal-rank fusion for \steering. The main comparison gives \steering this inference-time fusion, while \scoringmodel is reported without inference-time fusion.
\section{Experimental Setup}
\label{sec:setup}

\paragraph{Target models.}
We evaluate nine open-weight instruction-tuned LLMs: \texttt{TinyLlama-1.1B-Chat-v1.0}, \texttt{Llama-3.2-1B-Instruct}, \texttt{Qwen2-1.5B-Instruct}, \texttt{Llama-3.2-3B-Instruct}, \texttt{Qwen2.5-7B-Instruct}, \texttt{Llama-2-7b-chat-hf}, \texttt{Mistral-7B-Instruct-v0.3}, \texttt{Llama-3.1-8B-Instruct}, and \texttt{Qwen3-8B}.

\paragraph{Target-model continued pretraining.}
Before collecting responses, each target LLM is continued-pretrained on the \fakewiki text corpus with a causal language-modeling objective for 3 epochs at learning rate $2\times10^{-5}$. The training text consists of fabricated article text and document variants, not the QA probes or transformed queries. This exposure step is what makes the downstream task training-data attribution: the model responses are elicited from LLMs that have encountered the candidate documents as training text. The 80/20 document-ID split used later is for training and evaluating attribution methods, not for withholding documents from the target LLM.

\paragraph{Baselines.}
We compare against eleven retrieval baselines. MinHash estimates lexical resemblance by hashing token shingles and comparing compact signatures, giving a fast approximation to Jaccard overlap \citep{broder1997resemblance}; we run it over the answer alone and over the question-answer pair. The finetuned dense LLM embedding baseline uses an in-domain embedding head and ranks documents by cosine similarity. The remaining dense retrieval baselines use \sbert \citep{reimers2019sentence}, BGE \citep{xiao2024c}, and Contriever \citep{izacard2022contriever}: MiniLM, MPNet, BGE-base, and Contriever are evaluated with answer-only queries and with QA queries. Answer-only retrieval tests whether the generated response itself carries enough source evidence; QA retrieval tests a stronger setting where the retriever also sees the original question. In aggregate tables, ``best baseline'' means the strongest of these eleven baselines for that model and query condition.

\paragraph{Metrics and selection.}
The main paper reports Recall@10, with full per-model R@1 and R@5 tables in Appendix~\ref{app:per-model-r1-r5-tables}. For \scoringmodel, feature mode and checkpoint selection use held-out Clean validation data per target model, and the selected no-fusion scorer is evaluated unchanged on transformed queries. For \steering, validation selects the retrieval-fusion setting per target model and query condition. Further implementation details are in Appendix~\ref{app:impl}.

\section{Main Results}
\label{sec:results}

\subsection{Aggregate Recall@10 Results}

Table~\ref{tab:perjb} summarizes Recall@10 averaged across all nine target models. \scoringmodel improves over the strongest baseline in every query condition. The gains are largest for \texttt{RolePlay} and \texttt{NoiseInjection}, where surface retrieval remains plausible but unreliable, and smallest for \texttt{Obfuscate} and \texttt{Indirect}, which represent two different hard cases: shallow lexical substitution can sometimes favor response-only semantic retrieval, while indirect prompting reduces all methods to low absolute recall.

\begin{table}[t]
\centering
\caption{Recall@10 averaged across nine target LLMs. The best baseline is selected from eleven retrieval baselines separately for each model and query condition. Best per row is in \textbf{bold}, second-best is \underline{underlined}.}
\label{tab:perjb}
\small
\resizebox{0.9\linewidth}{!}{%
\begin{tabular}{lccccc}
\toprule
Query condition & Best baseline & \steering & \scoringmodel & $\Delta$ vs. baseline & $\Delta$ vs. \steering \\
\midrule
Clean          & 55.7 & \underline{69.2} & \textbf{77.2} & +21.5 & +8.0 \\
Obfuscate      & \underline{39.1} & 30.5 & \textbf{44.4} & +5.3  & +13.9 \\
RolePlay       & 42.9 & \underline{50.1} & \textbf{62.5} & +19.6 & +12.4 \\
NoiseInjection & 36.7 & \underline{47.0} & \textbf{59.2} & +22.5 & +12.2 \\
Indirect       & 12.0 & \underline{14.5} & \textbf{17.7} & +5.7  & +3.2 \\
\midrule
Average        & 37.3 & \underline{42.3} & \textbf{52.2} & +14.9 & +9.9 \\
\bottomrule
\end{tabular}%
}
\end{table}

Across all $9\times5$ cells, \scoringmodel beats the best baseline in 41/45 cells and beats \steering in 40/45 cells. Mean transformed-query Recall@10 improves by 13.2 points over the best baseline.
Additional aggregate summaries, including full win counts and transformed-query averages by target model, are reported in Appendix~\ref{app:aggregate}.

\paragraph{Seed robustness.}
We re-train each Clean-validation-selected \scoringmodel configuration with three seeds. The average per-cell standard deviation is 0.94 Recall@10 points, with variance largest on \texttt{Obfuscate}; full mean$\pm$std results are in Appendix~\ref{app:seed-robustness}.

\subsection{Per-Method Results on Large Models}

Table~\ref{tab:main-wide} reports Recall@10 for \texttt{Qwen3-8B} and \texttt{Llama-3.1-8B}, with methods as rows and query conditions as columns. These two models are useful stress cases because the best retrieval baseline varies across conditions: QA-style dense retrieval is strongest on clean prompts, while answer-only BGE or \sbert can become competitive under \texttt{Obfuscate}. Even against this condition-specific best-of-baselines comparison, \scoringmodel\ wins all ten large-model columns, while \steering is the strongest non-supervised method in most non-obfuscation columns. The remaining seven target-model tables are reported in Appendix~\ref{app:per-model-tables}, and the aggregate trends across all models are summarized above.

\begin{table*}[t]
\centering
\caption{Large-model Recall@10 table. Rows include all baselines and our attribution methods for two instruction-tuned target models. Best per column is in \textbf{bold}, second-best is \underline{underlined}.}
\label{tab:main-wide}
\setlength{\tabcolsep}{3.2pt}
\small
\resizebox{\textwidth}{!}{%
\begin{tabular}{lccccc ccccc}
\toprule
\multirow{2}{*}{Method} & \multicolumn{5}{c}{Qwen3-8B} & \multicolumn{5}{c}{Llama-3.1-8B} \\
\cmidrule(lr){2-6}\cmidrule(l){7-11}
& Clean & Obfuscate & RolePlay & NoiseInjection & Indirect & Clean & Obfuscate & RolePlay & NoiseInjection & Indirect \\
\midrule
MinHash (answer)       & 0.1 & 0.1 & 0.1 & 0.1 & 0.1 & 0.0 & 0.1 & 0.1 & 0.1 & 0.0 \\
MinHash (QA)           & 0.1 & 0.1 & 0.1 & 0.1 & 0.1 & 0.0 & 0.1 & 0.1 & 0.1 & 0.0 \\
Finetuned EmbedSim     & 17.4 & 9.4 & 16.1 & 16.4 & 2.3 & 16.9 & 15.9 & 23.6 & 23.1 & 2.2 \\
\sbert-MiniLM (answer) & 28.4 & 40.6 & 23.5 & 23.9 & 9.3 & 3.7 & 35.7 & 10.6 & 12.9 & 3.5 \\
\sbert-MPNet (answer)  & 25.5 & 36.3 & 20.3 & 20.1 & 8.0 & 3.6 & 32.4 & 9.7 & 11.6 & 3.0 \\
\sbert-MiniLM (QA)     & 39.7 & 12.4 & 29.1 & 25.0 & 9.1 & 28.9 & 9.5 & 19.9 & 18.5 & 4.0 \\
\sbert-MPNet (QA)      & 38.1 & 3.7 & 27.5 & 14.8 & 7.9 & 29.1 & 3.9 & 21.6 & 9.8 & 3.5 \\
BGE-base (answer)      & 30.8 & \underline{43.0} & 26.0 & 25.7 & 11.4 & 4.3 & 37.3 & 11.7 & 13.1 & 3.8 \\
BGE-base (QA)          & 43.1 & 16.6 & 40.3 & 23.6 & 11.3 & 33.3 & 13.4 & 30.6 & 13.8 & 5.9 \\
Contriever (answer)    & 32.4 & 42.8 & 26.2 & 26.6 & 10.6 & 3.6 & 27.9 & 11.3 & 12.5 & 2.7 \\
Contriever (QA)        & 30.1 & 22.8 & 22.3 & 15.0 & 3.5 & 5.1 & 10.8 & 9.3 & 7.1 & 0.8 \\
\midrule
\steering              & \underline{76.4} & 30.1 & \underline{60.5} & \underline{51.1} & \underline{20.1} & \underline{57.8} & \underline{37.7} & \underline{43.5} & \underline{38.1} & \underline{9.6} \\
\scoringmodel          & \textbf{78.1}{\scriptsize\,$\pm$0.5} & \textbf{53.9}{\scriptsize\,$\pm$1.5} & \textbf{63.8}{\scriptsize\,$\pm$0.7} & \textbf{62.1}{\scriptsize\,$\pm$0.3} & \textbf{21.6}{\scriptsize\,$\pm$0.8} & \textbf{76.7}{\scriptsize\,$\pm$0.2} & \textbf{59.5}{\scriptsize\,$\pm$2.1} & \textbf{63.4}{\scriptsize\,$\pm$0.6} & \textbf{63.5}{\scriptsize\,$\pm$0.4} & \textbf{18.2}{\scriptsize\,$\pm$0.4} \\
\bottomrule
\end{tabular}%
}
\end{table*}

The table shows that clean retrieval can be deceptively strong, but transformed prompts expose baseline instability: methods that work under clean prompting may fall sharply under indirect prompting or noise injection, and obfuscation can flip the strongest baseline from QA retrieval to answer-only retrieval. By contrast, \scoringmodel remains high across all five query conditions for both large models, including settings where the best baseline is below 15 Recall@10. The corresponding Recall@1 and Recall@5 tables in Appendix~\ref{app:per-model-r1-r5-tables} show that this advantage is not only a top-10 effect: \scoringmodel\ also wins every stricter-cutoff column for \texttt{Qwen3-8B} and \texttt{Llama-3.1-8B}.

\subsection{Scaling on Transformed Prompts}

\begin{figure}[t]
\centering
\begin{subfigure}[t]{0.48\linewidth}\centering
\includegraphics[width=\linewidth]{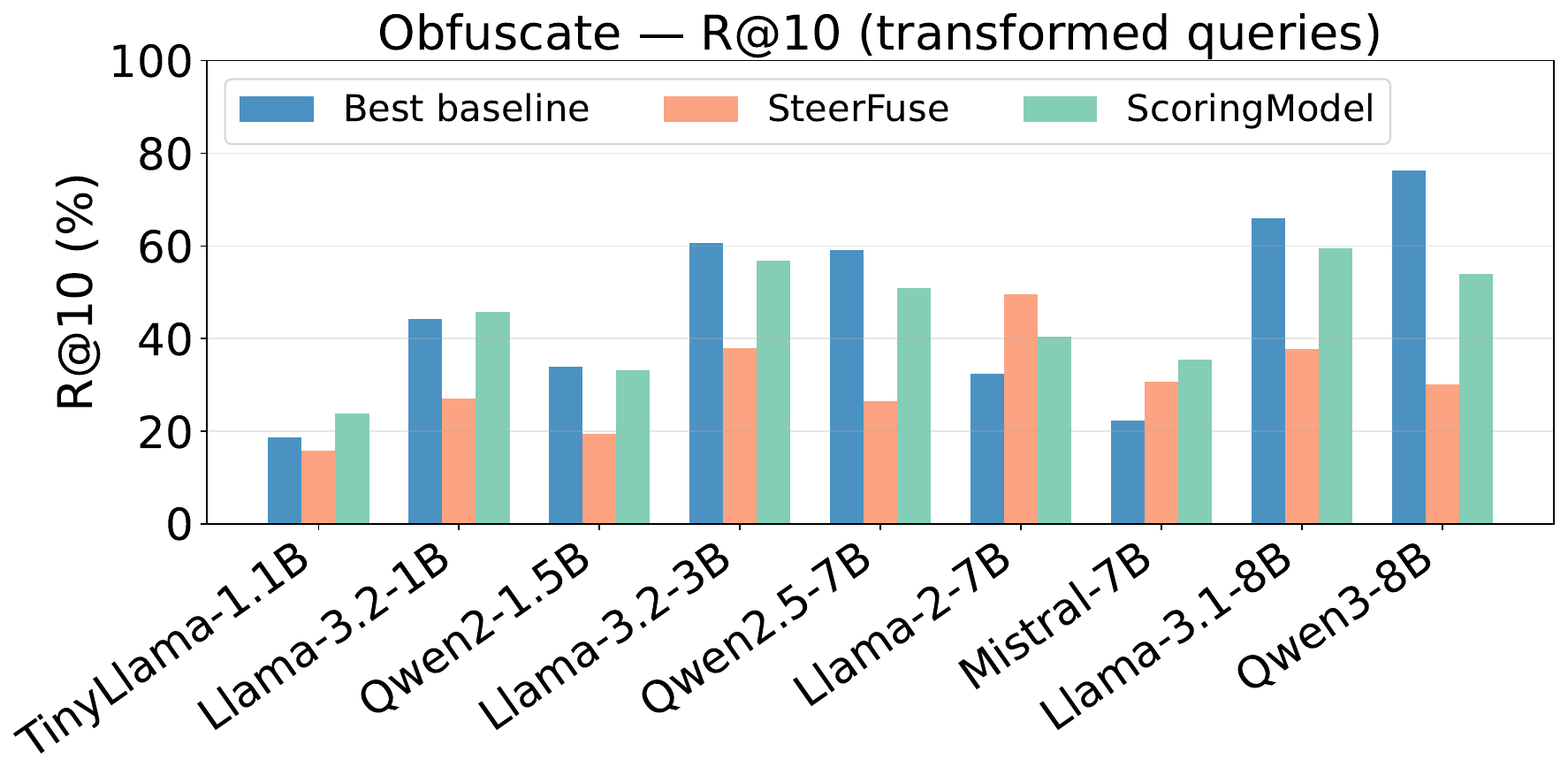}
\caption{Obfuscate}
\end{subfigure}\hfill
\begin{subfigure}[t]{0.48\linewidth}\centering
\includegraphics[width=\linewidth]{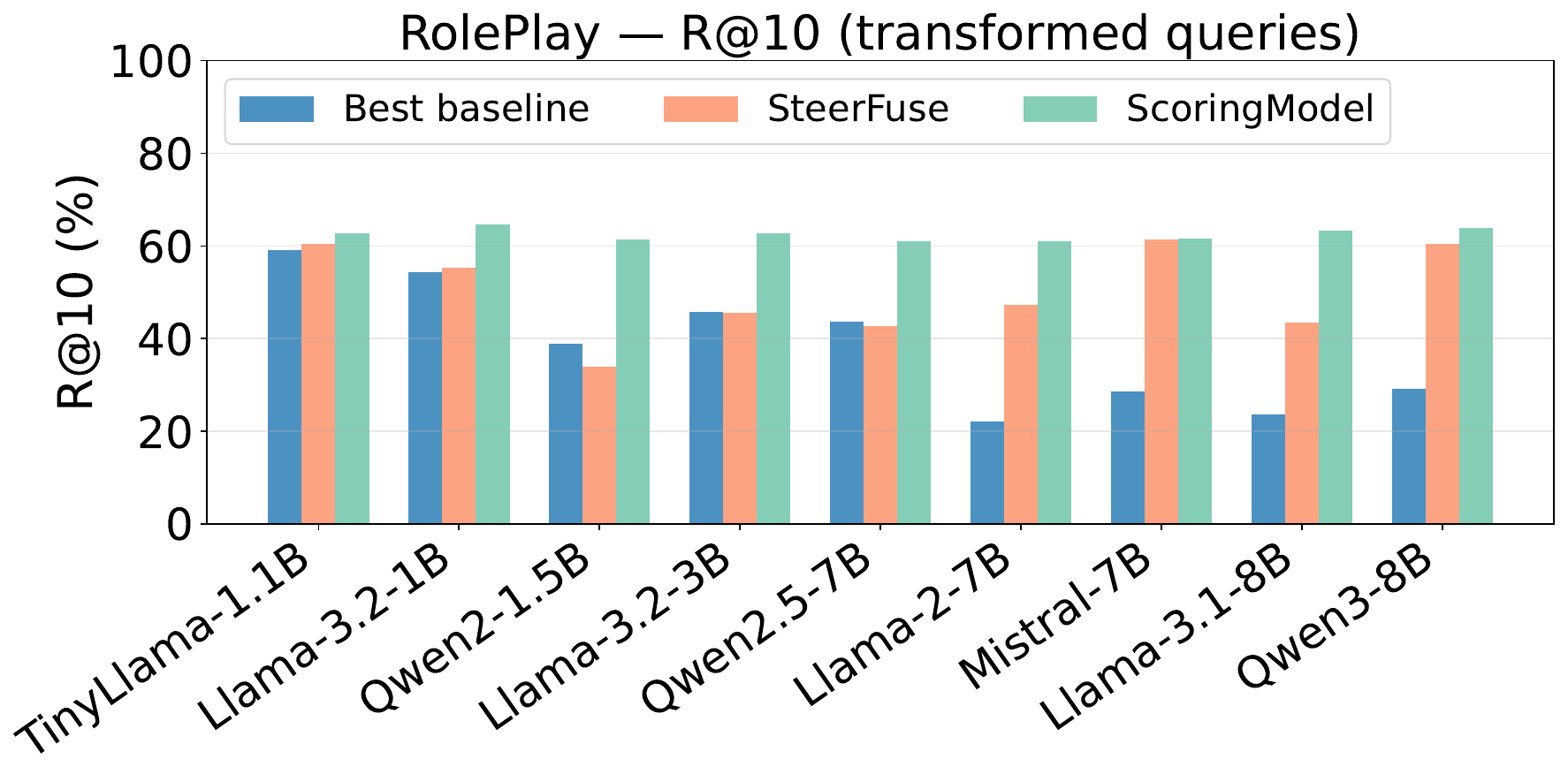}
\caption{RolePlay}
\end{subfigure}

\begin{subfigure}[t]{0.48\linewidth}\centering
\includegraphics[width=\linewidth]{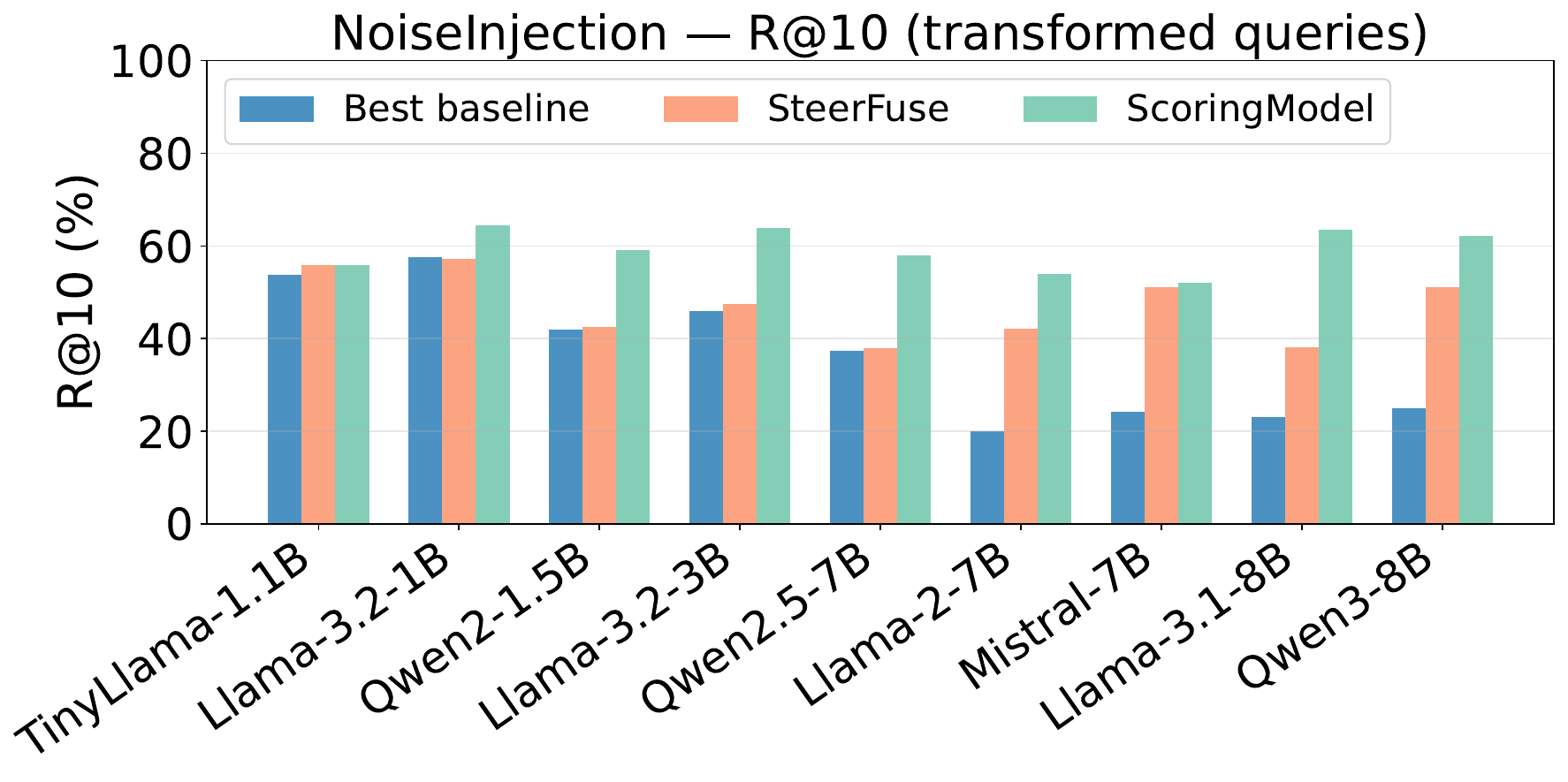}
\caption{NoiseInjection}
\end{subfigure}\hfill
\begin{subfigure}[t]{0.48\linewidth}\centering
\includegraphics[width=\linewidth]{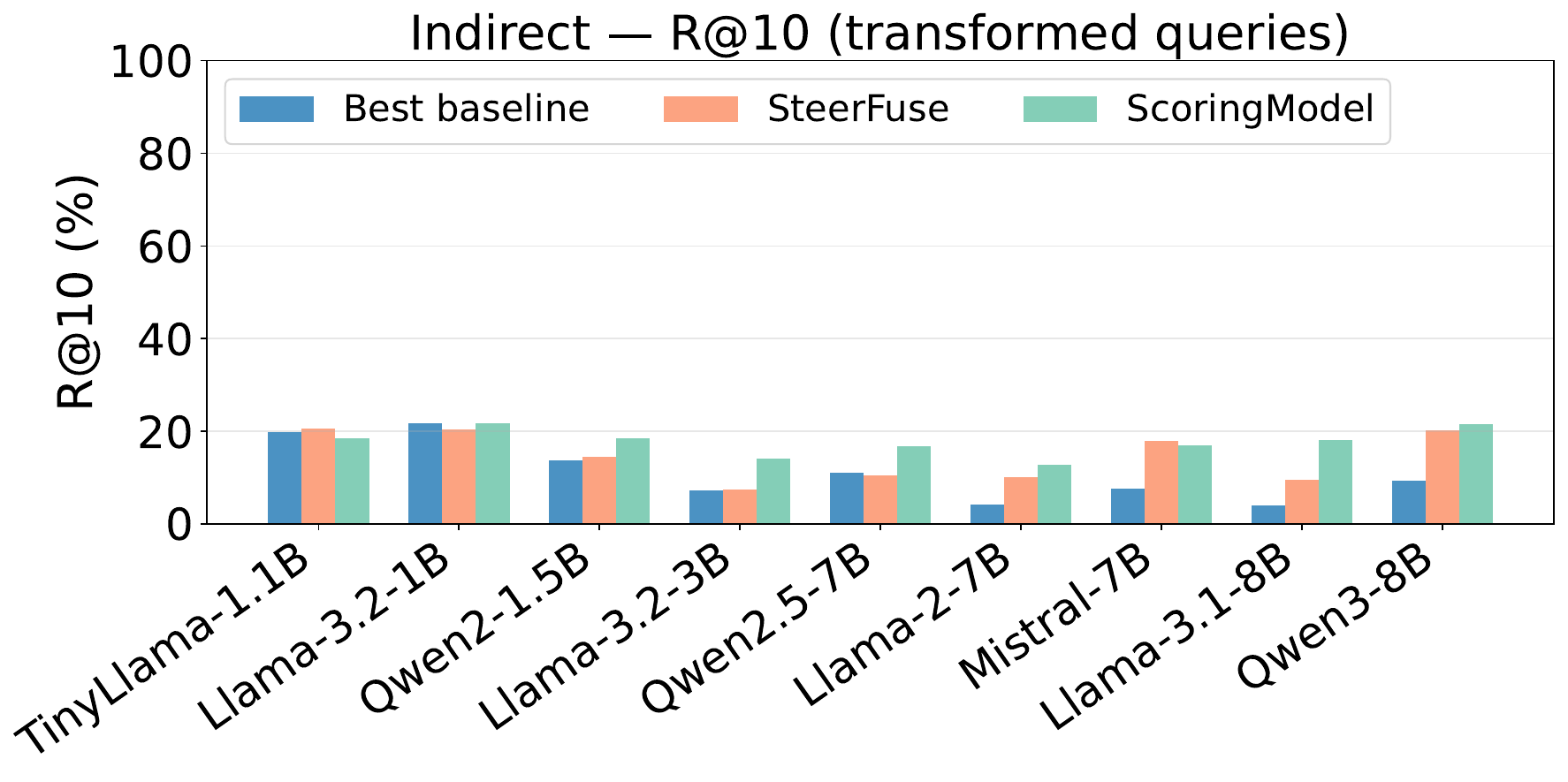}
\caption{Indirect}
\end{subfigure}
\caption{Improvement of \scoringmodel over the best baseline on transformed query conditions.}
\label{fig:transformed}
\end{figure}

Figure~\ref{fig:transformed} visualizes the per-model improvement of \scoringmodel over the best baseline on the four transformed query conditions. The largest gains appear for the larger target models: \texttt{Llama-3.1-8B} improves by +26.9 Recall@10 on average over transformed conditions, \texttt{Qwen3-8B} by +20.0, \texttt{Llama-2-7B} by +19.0, and \texttt{Mistral-7B} by +17.0. Smaller models still benefit, but with narrower margins. This pattern suggests that larger LLM hidden states encode more recoverable provenance information, making attribution easier for a trained scorer even as the underlying model is larger.

\subsection{Ablation Study}
\label{sec:ablation-brief}

Appendix~\ref{app:ablations} provides full ablations. The main takeaway is that no-fusion \scoringmodel is already a strong standalone scorer: it uses a single learned compatibility score, requires no test-time mixing weight, and wins 41/45 model-by-condition cells. We evaluate \scoringmodel--\sbert fusion only as an ablation.

\steering shows a different pattern: its mean gain over the stronger of activation-only and \sbert-only rankings is only +2.0 Recall@10, and about 96\% of the fusion uplift comes from \sbert. The main exception is \texttt{Obfuscate}, where the activation component contributes about +7.5 points, suggesting that internal-state evidence helps most when lexical cues are actively distorted.

Other ablations show that the conclusions are not driven by a single representation or combiner: z-score and reciprocal-rank fusion give similar trends, and feature-mode diagnostics show complementary hidden-state and QA-style signals.

\section{Conclusion}

We presented \method, a benchmark and method suite for robust pinpoint provenance in LLM responses. \fakewiki shows that clean retrieval can overstate attribution reliability: under obfuscation, role-play, noise, and indirect prompts, methods must distinguish true answer support from topical resemblance. This distinction matters for audit workflows because a high-similarity document is not necessarily the document that supplied the answer-critical fact. In this setting, \scoringmodel improves mean Recall@10 from 37.3 to 52.2 and wins 41/45 model-by-condition cells, with the same trend holding at stricter Recall@1 and Recall@5 cutoffs. The gains are largest on transformed prompts and larger target models, suggesting that supervised attribution can recover provenance signals that generic retrievers miss when surface cues are unstable. \steering shows that activation-space evidence is useful when stabilized by text retrieval, but is less uniform than the supervised scorer; this points to internal-state evidence as a promising complement rather than a complete replacement for retrieval.

Overall, robust training data attribution should be evaluated with hard negatives and prompt transformations that expose failures hidden by clean semantic retrieval. We view pinpoint provenance systems as audit aids that return inspectable candidate sources, and future work should push beyond document-level recall toward calibrated confidence and finer-grained evidence localization.

\bibliographystyle{plainnat}
\bibliography{reference}

\appendix

\section{Limitations and Future Work}
\label{app:limitations-first}

\paragraph{Limitations.}
Our study provides a controlled and fine-grained evaluation of pinpoint provenance, with ground-truth source documents, hard negatives, source-preserving variants, and transformed query conditions that stress different attribution failure modes. However, several aspects could be further improved. First, our transformed query conditions are controlled stress tests rather than a full taxonomy of user prompting behavior. They are designed to isolate different shortcut pressures, not to exhaust every possible way a prompt can alter the surface form of a response. Second, our main metric is Recall@10, which is appropriate for audit workflows where a human inspector reviews a short candidate list, but it does not capture all downstream notions of attribution usefulness, such as calibrated confidence, explanation quality, or evidence localization within a document. Finally, our evaluation focuses on document-level attribution. This is useful for source inspection, but it does not identify the exact sentence or span that supports a response.

\paragraph{Future work.}
Future work can address these limitations in three directions. First, the transformed-query suite can be expanded to cover a broader range of prompt styles, including more natural user rewrites, multi-turn interactions, and compositional transformations that combine obfuscation, role-play, noise, and indirect prompting. Second, evaluation can move beyond Recall@10 by adding calibration and ranking-quality metrics, so that provenance systems report not only whether a true source appears in the top candidates, but also how reliable the returned evidence is. Third, future benchmarks can extend document-level attribution to finer-grained evidence localization, such as sentence-level or span-level provenance, which would make the returned results easier for human auditors to verify.

\section{Broader Impacts}
\label{app:broader-impacts}

This work is intended to support more reliable auditing of language-model outputs by helping identify candidate source documents that may support a generated response. Potential positive impacts include improved dataset curation, copyright and provenance analysis, misinformation forensics, and safety debugging. At the same time, provenance scores should be interpreted carefully: a high-ranked document is evidence of source support within the candidate corpus, not a definitive causal claim about model training or generation. Misinterpreting attribution results could lead to overconfident conclusions about whether a particular document caused an output. We therefore frame \method as an audit aid that returns inspectable evidence for human review, rather than as an automated system for assigning legal or causal responsibility.

\section{Asset Licenses}
\label{app:asset-licenses}

We use publicly available models and software tools, including the open-weight target LLMs listed in Section~\ref{sec:setup}, Sentence-BERT embedding models \citep{reimers2019sentence}, BGE embeddings \citep{xiao2024c}, Contriever embeddings \citep{izacard2022contriever}, PyTorch, CUDA, and standard Python libraries. We cite the corresponding model families, methods, and software tools where appropriate, and use these assets for research evaluation under their respective licenses and terms of use. The newly introduced \fakewiki benchmark consists of fabricated documents generated for this study, and the anonymized release includes the benchmark data, prompts, and evaluation code.

\section{Activation-Steering Approximation Details}
\label{app:steering-approx}

This appendix expands the intervention view of \steering from Section~\ref{sec:steering}. The method starts from an exact but expensive activation-patching score, then uses a sequence of approximations: a first-order Taylor approximation that turns patched forward passes into dot products, and an LM-head approximation that avoids a backward pass for the sensitivity vector by replacing the gradient direction with answer-token LM-head rows.

\subsection{Exact Activation-Patching Score}

Fix a target response $y_{1:m}$ and a candidate document direction $\vv_D$. At the selected layer $\ell^\star$, let $\vh^{(\ell^\star)}_{t_i}$ be the hidden state used to predict answer token $y_i$. The exact intervention asks whether mixing this candidate direction into the answer-token hidden states increases the model probability of the observed response:
\begin{equation}
\tilde{\vh}^{(D)}_{t_i}(\alpha)
=(1-\alpha)\vh^{(\ell^\star)}_{t_i}+\alpha\vv_D,
\qquad \alpha\in(0,1].
\label{eq:appendix-convex-patch}
\end{equation}
The corresponding exact score is
\begin{equation}
\Delta_{\mathrm{exact}}(D)
=\sum_{i=1}^{m}\left[
\log p_\theta(y_i\mid y_{<i};\tilde{\vh}^{(D)}_{t_i}(\alpha))
-\log p_\theta(y_i\mid y_{<i};\vh^{(\ell^\star)}_{t_i})
\right].
\label{eq:appendix-exact-gain}
\end{equation}
A large positive value means that injecting document $D$ makes the target model more confident in the fixed response. This is the most direct causal-style score, but evaluating it for $N$ documents requires one baseline forward pass plus $N$ patched forward passes.

\subsection{First-Order Taylor Approximation}

To avoid a patched forward pass for every candidate document, define the scalar objective
\begin{equation}
f(\{\vh_{t_i}\}_{i=1}^{m})
=\sum_{i=1}^{m}\log p_\theta(y_i\mid y_{<i}),
\label{eq:appendix-answer-logprob}
\end{equation}
the log-probability of the whole fixed answer under the unpatched hidden states. A first-order Taylor expansion around the original hidden states gives
\begin{align}
f(\tilde{\vh}^{(D)})-f(\vh)
&=\alpha\sum_{i=1}^{m}
\underbrace{\nabla_{\vh^{(\ell^\star)}_{t_i}}f(\vh)^{\top}}_{\vg_i^{\top}}
(\vv_D-\vh^{(\ell^\star)}_{t_i})+O(\alpha^2) \\
&=\alpha\left(\sum_{i=1}^{m}\vg_i\right)^{\top}\vv_D
-\alpha\sum_{i=1}^{m}\vg_i^{\top}\vh^{(\ell^\star)}_{t_i}
+O(\alpha^2).
\label{eq:appendix-taylor-expanded}
\end{align}
The second term does not depend on $D$, so it does not affect the ranking of candidate documents. If
\begin{equation}
\vg=\sum_{i=1}^{m}\vg_i
=\sum_{i=1}^{m}
\frac{\partial}{\partial \vh^{(\ell^\star)}_{t_i}}
\log p_\theta(y_i\mid y_{<i}),
\label{eq:appendix-sensitivity}
\end{equation}
then the document-dependent part of the first-order score is
\begin{equation}
\widehat{\Delta\log p}(D)\propto \vg^{\top}\vv_D.
\label{eq:appendix-gradient-score}
\end{equation}
This approximation turns document ranking into a matrix-vector product once all document vectors have been cached. It is accurate when the mixing weight is small enough that higher-order terms in $\alpha$ do not dominate the ranking.

\subsection{Approximating Sensitivity Without Backpropagation}

Computing $\vg$ exactly requires one backward pass for each response. This is much cheaper than $N$ patched forward passes, but it can still be expensive when many responses must be attributed. When the patched layer is the final hidden layer before the language-model head, the per-token sensitivity can be written in closed form. Let $\mW\in\R^{|V|\times d}$ be the LM-head matrix and let $\mW_w$ denote the row associated with vocabulary token $w$. For the hidden state $\vh$ at an answer position, the logits are
\begin{equation}
z_w=\vh^{\top}\mW_w,
\qquad
p_\theta(w\mid \mathrm{ctx})=\frac{\exp(\vh^{\top}\mW_w)}{\sum_{u\in V}\exp(\vh^{\top}\mW_u)}.
\end{equation}
Therefore, for answer token $y_i$,
\begin{align}
\vg_i
&=\frac{\partial}{\partial\vh}\log p_\theta(y_i\mid y_{<i}) \\
&=\mW_{y_i}-\sum_{w\in V}p_\theta(w\mid y_{<i})\mW_w \\
&=\mW_{y_i}-\mathbb{E}_{w\sim p_\theta(\cdot\mid y_{<i})}[\mW_w].
\label{eq:appendix-lm-head-gradient}
\end{align}
This formula shows that the sensitivity direction points toward the LM-head row of the observed answer token and away from the probability-weighted average LM-head direction. Computing the expectation exactly requires multiplying the full vocabulary probability vector by $\mW$, which can be costly. A further approximation drops this expectation term and uses
\begin{equation}
\tilde{\vg}_i\propto \mW_{y_i},
\qquad
\tilde{\vg}=\sum_{i=1}^{m}\tilde{\vg}_i.
\label{eq:appendix-head-row-proxy}
\end{equation}
This is reasonable when the answer token direction dominates the local gradient, but it is not an exact identity: the omitted expectation term can matter when the next-token distribution is diffuse or when competing tokens have substantial probability mass. Under this approximation, the Taylor score becomes
\begin{equation}
\widehat{\Delta\log p}(D)\propto \tilde{\vg}^{\top}\vv_D,
\end{equation}
so ranking still reduces to a dot product, but the response-side vector can be obtained by LM-head row lookups rather than backpropagation.

\subsection{Implemented LM-Head Row Proxy}

The gradient score in Eq.~\eqref{eq:appendix-gradient-score} removes the need for one patched forward pass per document, and the LM-head approximation in Eq.~\eqref{eq:appendix-head-row-proxy} removes the backward pass. The implemented \steering method represents the answer by the normalized sum of its generated-token LM-head rows and ranks documents by cosine similarity against cached document directions:
\begin{equation}
s_{\mathrm{act}}(r,D)=\cos\left(
\sum_{i=1}^{m}\mW_{y_i},\vv_D
\right).
\end{equation}
In experiments, we use a finer-grained version of the same score: long documents are split into sentence-respecting chunks, each chunk receives a score, and the document score is the maximum over its chunks. This proxy is cheaper than exact patching and avoids backpropagation, but it is still noisy because LM-head rows are only an approximation to the full gradient sensitivity in Eq.~\eqref{eq:appendix-lm-head-gradient}. For this reason, the main \steering method fuses the activation score with \sbert-QA retrieval rather than using the activation score alone.

\section{Additional Aggregate Results}
\label{app:aggregate}

Table~\ref{tab:winrate} reports head-to-head win counts over all 45 model-by-query-condition cells. The pattern is consistent with the main aggregate results: \scoringmodel is not only better on average, but also wins across most individual settings. It outperforms the strongest retrieval baseline in 41/45 cells and outperforms \steering in 40/45 cells. At the same time, \steering beats the best baseline in 32/45 cells, suggesting that activation-based evidence can be useful when stabilized by retrieval fusion.

\begin{table}[h]
\centering
\caption{Head-to-head win counts over all 45 model-by-query-condition cells.}
\label{tab:winrate}
\small
\begin{tabular}{lcc}
\toprule
Comparison & Wins / 45 & Percent \\
\midrule
\scoringmodel $>$ best baseline & 41 & 91\% \\
\scoringmodel $>$ \steering & 40 & 89\% \\
\steering $>$ best baseline & 32 & 71\% \\
\bottomrule
\end{tabular}
\end{table}

Table~\ref{tab:model-avg} averages Recall@10 across the four transformed query conditions, excluding clean prompts. This table highlights where the robustness gains are largest. The biggest improvements occur for Llama-3.1-8B and Qwen3-8B, where \scoringmodel improves over the best baseline by +26.9 and +20.0 points respectively. The next strongest gains appear for Llama-2-7B and Mistral-7B. Smaller models still benefit, but with narrower margins. This supports the main-paper observation that larger target models tend to expose more recoverable provenance signal for a supervised scorer, especially when surface-form cues are disrupted.

\begin{table}[h]
\centering
\caption{Average transformed-query Recall@10 by target model. Clean prompts are excluded.}
\label{tab:model-avg}
\small
\begin{tabular}{lcccc}
\toprule
Model & Best baseline & \steering & \scoringmodel & $\Delta$ vs. baseline \\
\midrule
Llama-3.1-8B & 24.2 & 32.2 & \textbf{51.1} & +26.9 \\
Qwen3-8B & 30.3 & 40.4 & \textbf{50.4} & +20.0 \\
Llama-2-7B & 23.0 & 37.3 & \textbf{42.0} & +19.0 \\
Mistral-7B & 24.6 & 40.2 & \textbf{41.5} & +17.0 \\
Qwen2-1.5B & 32.1 & 27.6 & \textbf{43.0} & +10.9 \\
Llama-3.2-3B & 39.9 & 34.6 & \textbf{49.4} & +9.5 \\
Qwen2.5-7B & 37.9 & 29.4 & \textbf{46.6} & +8.8 \\
Llama-3.2-1B & 44.4 & 40.0 & \textbf{49.2} & +4.8 \\
TinyLlama-1.1B & 37.9 & 38.2 & \textbf{40.3} & +2.4 \\
\bottomrule
\end{tabular}
\end{table}

Overall, these aggregate results show that the gains are not driven by a small number of favorable cases. \scoringmodel is consistently strong across both head-to-head comparisons and transformed-query averages, while \steering provides a useful but less reliable intermediate signal.

\section{Ablation Details}
\label{app:ablations}

\paragraph{Inference-time fusion.}
Adding \sbert fusion to \scoringmodel improves mean Recall@10 by +5.5 points over the stronger of no-fusion \scoringmodel and \sbert alone. The gains are distributed across query conditions: Clean +2.7, Obfuscate +5.5, RolePlay +9.5, NoiseInjection +8.6, and Indirect +1.0. We keep this variant as an ablation because the no-fusion method is simpler, requires no test-time fusion weight, and already wins the main comparison.

\paragraph{\steering decomposition.}
For \steering, the mean fusion gain over the stronger of activation-only and \sbert-only rankings is +2.0 Recall@10, and approximately 96\% of the fusion uplift comes from the \sbert component. The main exception is \texttt{Obfuscate}, where the activation component contributes about +7.5 points. Thus activation steering is informative, but not yet a strong standalone provenance ranker.

\paragraph{Combiner choice.}
Z-score fusion wins the majority of \scoringmodel-fusion cells and \steering cells, so the main paper uses z-score when reporting \steering. Reciprocal-rank fusion is reported as a sensitivity check and is most useful under \texttt{Indirect}, where score distributions shift strongly.

\begin{figure}[h]
\centering
\begin{subfigure}[t]{0.45\linewidth}\centering
\includegraphics[width=\linewidth]{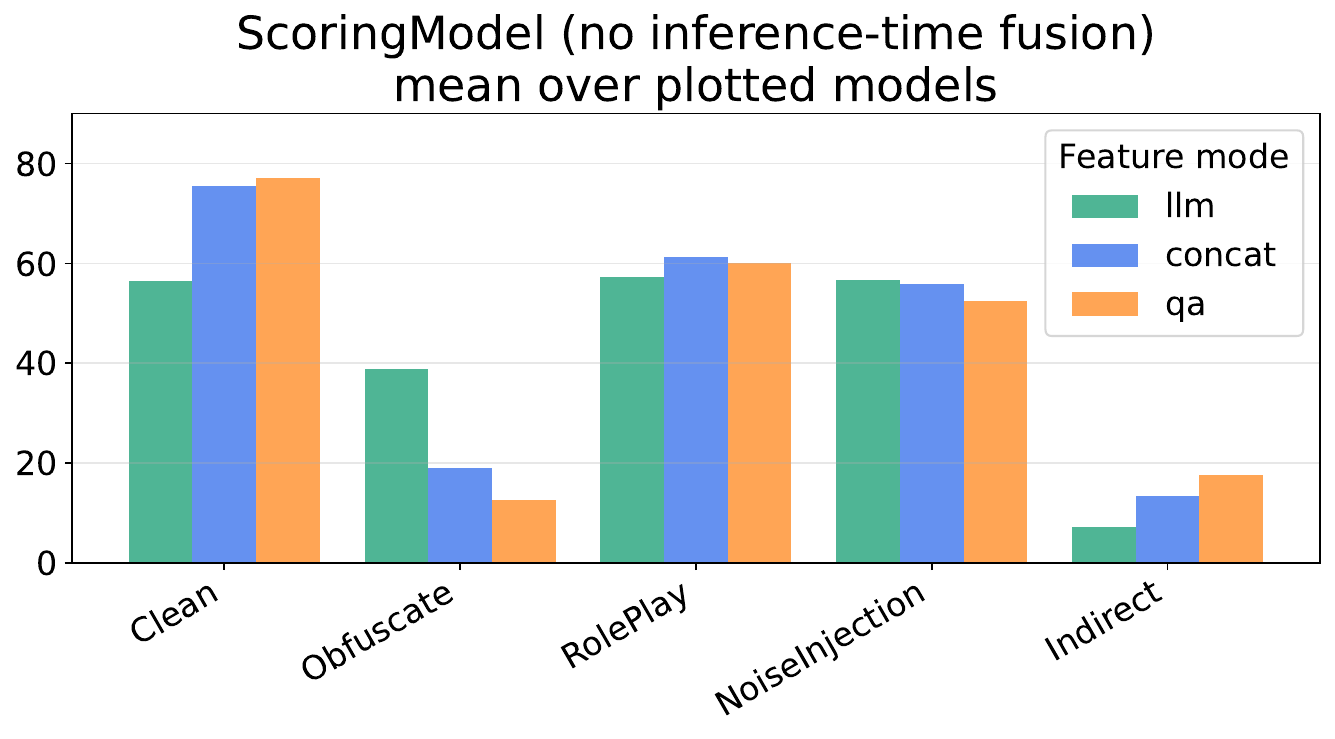}
\caption{No-fusion \scoringmodel}
\end{subfigure}\hfill
\begin{subfigure}[t]{0.45\linewidth}\centering
\includegraphics[width=\linewidth]{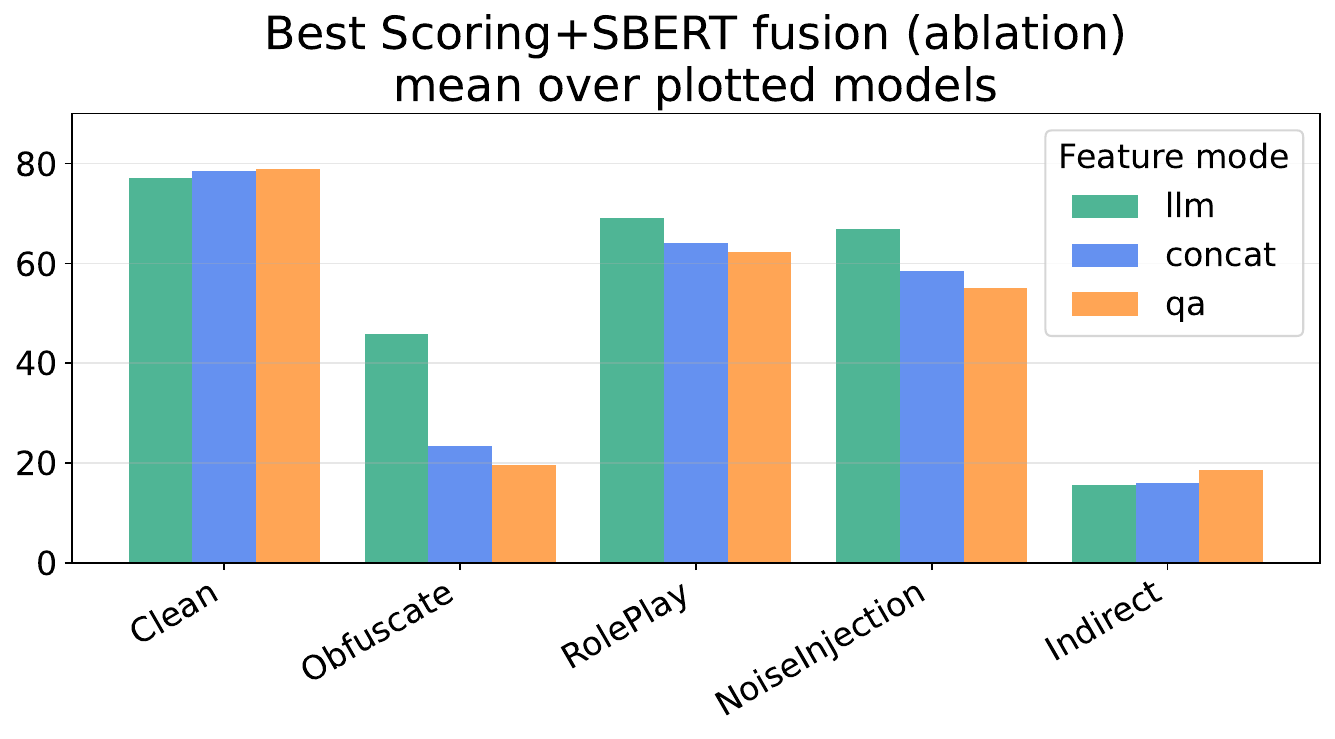}
\caption{Fusion ablation}
\end{subfigure}
\caption{Feature-mode ablations for \scoringmodel. Left: no inference-time fusion. Right: best \scoringmodel+\sbert fusion. Hidden-state features are most useful under obfuscation and, with fusion, become the strongest feature mode on most transformed conditions.}
\label{fig:feature-mode-ablation}
\end{figure}

\paragraph{Feature mode.}
We consider three input-feature variants for the \scoringmodel scorer. The \texttt{llm} feature is a mean-pooled hidden state from a selected layer of the target LLM, computed for the response side and for the candidate document side; it is intended to capture model-internal content representations that may survive obfuscation or paraphrase. The \texttt{qa} feature is a \sbert-MiniLM embedding \citep{reimers2019sentence} of a QA-style textual representation, which is strong when the question--answer pair retains semantic alignment with the source. The \texttt{concat} feature concatenates the two, allowing early fusion at the MLP input. We emphasize that all three variants are evaluated at $\lambda=0$: the MLP produces a single learned score per candidate, and there is no test-time score-level mixing with a separate \sbert ranker. Main results use the feature mode selected on Clean validation for each target model. The condition-wise feature results in Figure~\ref{fig:feature-mode-ablation} are diagnostic rather than a transformed-condition tuning procedure.

\paragraph{Grid size and embedding backbone.}
A wider negative-mining grid is sufficient for the \texttt{concat} setting; larger grids help \texttt{llm} features, especially on \texttt{Obfuscate}. MiniLM is stronger than MPNet on this benchmark across the relevant sweeps.

The figures below visualize the same ablation questions: which input
representation is most useful for \scoringmodel, whether z-score fusion or
reciprocal-rank fusion is the more important design choice, and how much the
validation-selected fusion models rely on the learned attribution signal versus
\sbert. All plots report Recall@10 and average over the nine target models
unless otherwise stated.

\paragraph{Feature representation.}
Figure~\ref{fig:feature-mode-ablation} compares three feature modes for
\scoringmodel. The \texttt{qa} feature is strongest on clean prompts in the
no-fusion setting, reaching roughly the high-70s Recall@10, because the
question-answer text remains semantically aligned with the source document. The
same text channel is also the best no-fusion feature on \texttt{Indirect}, where
all methods are low but QA semantics still preserve the most usable retrieval
signal. By contrast, the \texttt{llm} feature is clearly strongest on
\texttt{Obfuscate}: its average Recall@10 is roughly twice that of
\texttt{concat} and more than three times that of \texttt{qa}. This supports the
interpretation that target-model hidden states carry provenance information that
is less tied to surface word choice. For \texttt{RolePlay} and
\texttt{NoiseInjection}, the differences are smaller: \texttt{concat} is best on
role-play without fusion, while \texttt{llm} and \texttt{concat} are close on
noise injection. This is consistent with these transformations preserving more
of the original semantics while adding stylistic or irrelevant context.

The fusion ablation in Figure~\ref{fig:feature-mode-ablation} changes the
picture in an informative way. Once \sbert score-level fusion is allowed,
\texttt{llm} becomes the best feature mode for \texttt{Obfuscate},
\texttt{RolePlay}, and \texttt{NoiseInjection}. The text-only \texttt{qa} mode
remains competitive on clean prompts and remains best on \texttt{Indirect}, but
it no longer dominates the transformed settings. Thus the best fusion behavior
does not come from replacing hidden-state features with text retrieval; it comes
from using \sbert as a stabilizing semantic prior while letting model-internal
features contribute complementary signal under transformed prompts. This is why
the main paper reports the simpler no-fusion \scoringmodel as the primary method
and treats \scoringmodel+\sbert as an ablation: fusion can improve robustness,
but the learned scorer already provides a strong standalone attribution signal.

\begin{figure}[h]
\centering
\begin{subfigure}[t]{0.45\linewidth}\centering
\includegraphics[width=\linewidth]{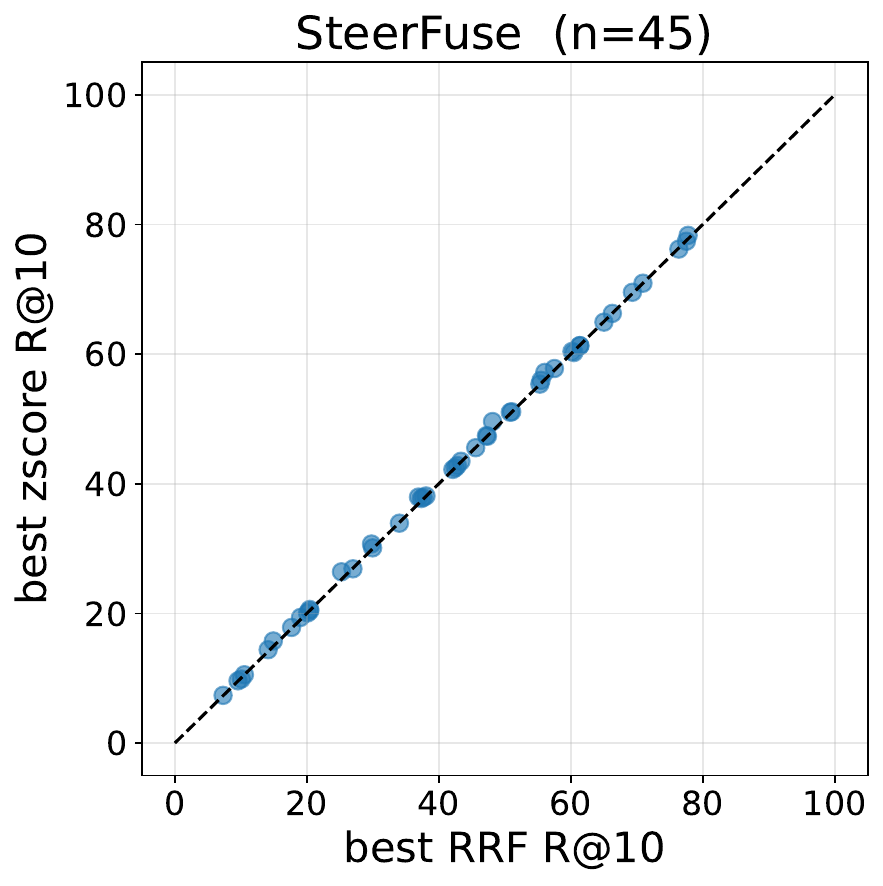}
\caption{\steering}
\end{subfigure}\hfill
\begin{subfigure}[t]{0.45\linewidth}\centering
\includegraphics[width=\linewidth]{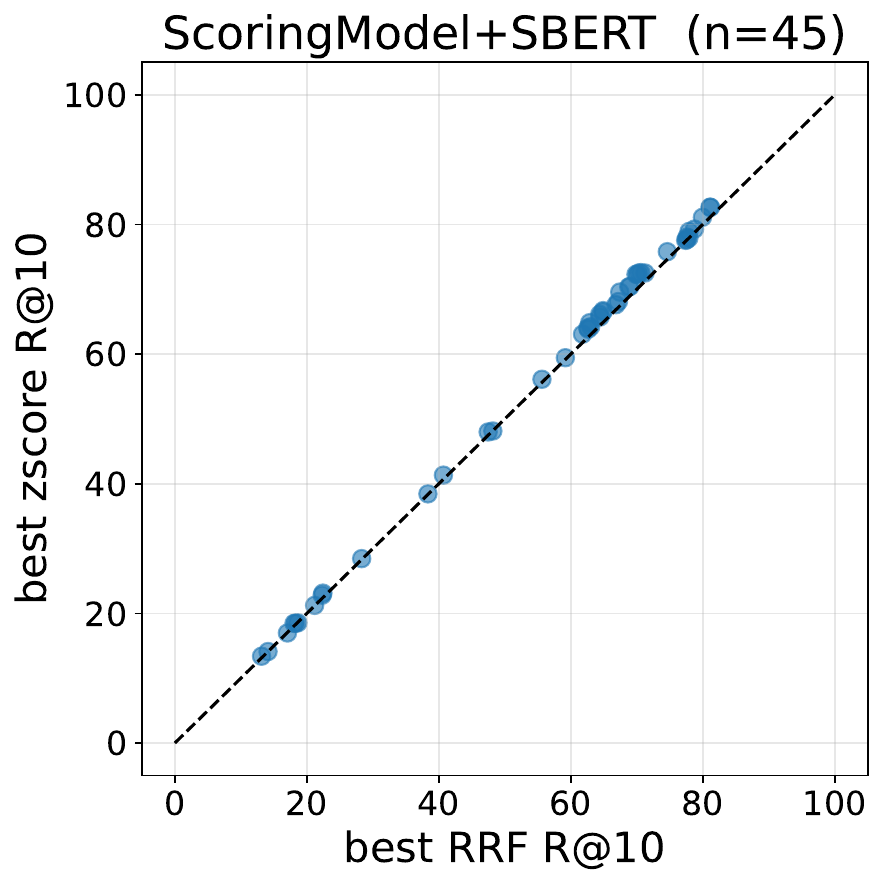}
\caption{\scoringmodel fusion ablation}
\end{subfigure}
\caption{Z-score versus reciprocal-rank fusion over all 45 model-by-condition cells. Points near the diagonal indicate that the fusion gains are not sensitive to the particular combiner.}
\label{fig:combiner-ablation}
\end{figure}

\paragraph{Fusion combiner.}
Figure~\ref{fig:combiner-ablation} compares the two score-combination rules used
in the fusion sweeps: z-score fusion and reciprocal-rank fusion (RRF). Each point
is one model-by-query-condition cell, so each panel contains 45 points. In both
the \steering panel and the \scoringmodel+\sbert panel, almost all points lie on
or very near the diagonal. This means that the choice of combiner is not the
main driver of performance: if a cell is easy or hard for fusion, it is usually
easy or hard under both normalization schemes. The practical implication is that
the results are not an artifact of a fragile score calibration trick. Z-score
fusion is slightly more favorable in some high-recall \scoringmodel cells, while
RRF is occasionally competitive when raw score scales are less comparable, but
the two views tell the same qualitative story.

\begin{figure}[h]
\centering
\begin{subfigure}[t]{0.45\linewidth}\centering
\includegraphics[width=\linewidth]{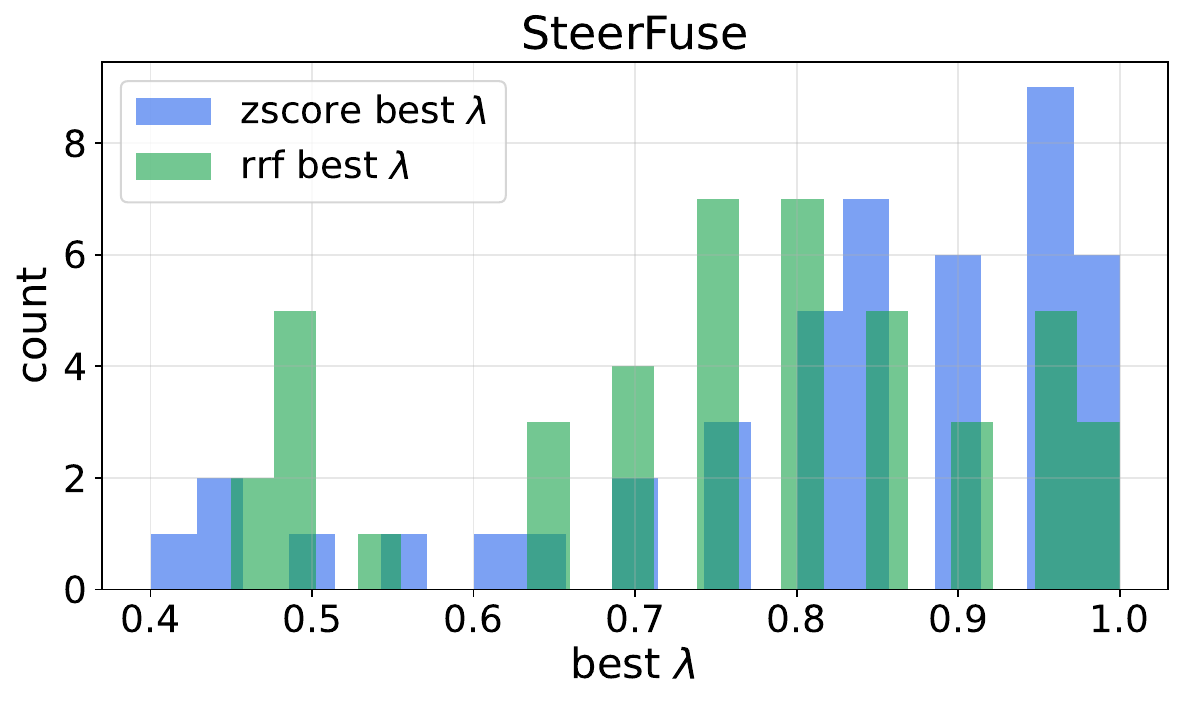}
\caption{\steering}
\end{subfigure}\hfill
\begin{subfigure}[t]{0.45\linewidth}\centering
\includegraphics[width=\linewidth]{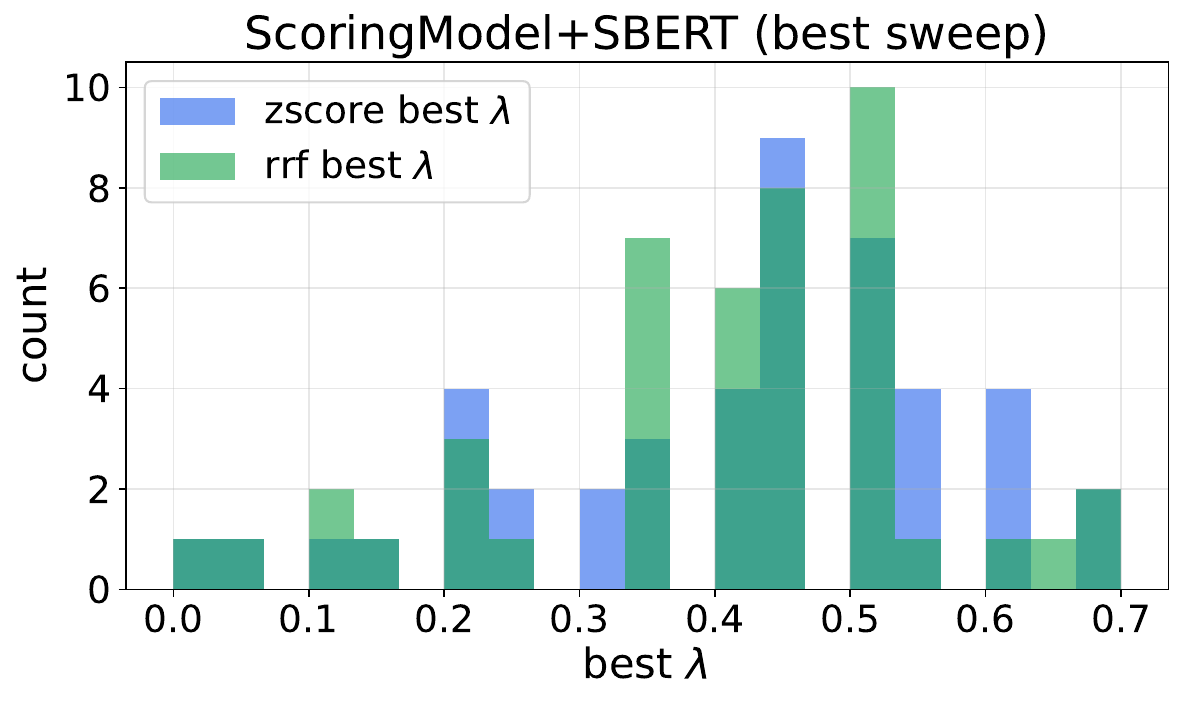}
\caption{\scoringmodel fusion ablation}
\end{subfigure}
\caption{Validation-selected fusion weights. $\lambda=0$ is method-only and $\lambda=1$ is \sbert-only. \steering selects large $\lambda$ values, while \scoringmodel+\sbert selects intermediate values, indicating more genuine complementarity.}
\label{fig:lambda-ablation}
\end{figure}

\paragraph{Fusion weights.}
Figure~\ref{fig:lambda-ablation} shows the validation-selected mixing weights.
Here $\lambda=0$ means the method-only score is used, $\lambda=1$ means
\sbert-QA alone is used, and intermediate values indicate genuine score-level
fusion. The \steering histogram is concentrated toward large $\lambda$ values,
mostly around $0.75$--$1.0$, with no mass near small $\lambda$. This confirms
that raw activation steering is a useful but noisy signal: validation usually
prefers to keep a large \sbert component and add steering only as a correction.
The \scoringmodel+\sbert histogram is different. Its selected weights are mostly
intermediate, with substantial mass around $0.35$--$0.55$ and little mass near
$\lambda=1$. This indicates that the learned \scoringmodel signal remains central
even when \sbert fusion is permitted; validation rarely chooses to discard the
learned scorer in favor of \sbert alone. The contrast between the two histograms
is therefore important: \steering fusion is largely \sbert-anchored, whereas
\scoringmodel fusion is closer to true complementarity between a learned
attribution scorer and a semantic retrieval prior.

\section{Seed Robustness}
\label{app:seed-robustness}

We report seed robustness for \scoringmodel because it is the only learned method with stochastic training; \steering is deterministic given the target-model forward passes, cached activations, and validation-selected fusion setting. Stability of \steering is instead assessed through the fusion-combiner and fusion-weight ablations in Appendix~\ref{app:ablations}.

For each target model we re-train the Clean-validation-selected \scoringmodel configuration under three independent seeds (42, 123, 2024) and report Recall@10 mean with standard deviation shown as a subscript per (model, query condition). Standard deviations are typically below 1.5 Recall@10 points; the larger values on \texttt{Obfuscate} reflect higher sensitivity of the synonym-substitution condition to negative-mining randomness.

\begin{table}[h]
\centering
\caption{\scoringmodel Recall@10 mean with standard deviation shown as a subscript across three seeds per (model, query condition).}
\label{tab:seed-robustness}
\small
\setlength{\tabcolsep}{4pt}
\begin{tabular}{lccccc}
\toprule
Model & Clean & Obfuscate & RolePlay & NoiseInjection & Indirect \\
\midrule
TinyLlama-1.1B   & 78.2$_{\pm 0.2}$ & 23.8$_{\pm 1.2}$ & 62.7$_{\pm 1.1}$ & 55.9$_{\pm 1.0}$ & 18.6$_{\pm 0.3}$ \\
Llama-3.2-1B     & 76.5$_{\pm 0.5}$ & 45.8$_{\pm 2.6}$ & 64.7$_{\pm 0.3}$ & 64.5$_{\pm 1.0}$ & 21.8$_{\pm 0.7}$ \\
Qwen2-1.5B       & 77.2$_{\pm 0.4}$ & 33.2$_{\pm 4.4}$ & 61.4$_{\pm 1.4}$ & 59.1$_{\pm 1.0}$ & 18.4$_{\pm 0.9}$ \\
Llama-3.2-3B     & 76.5$_{\pm 0.5}$ & 56.8$_{\pm 4.3}$ & 62.8$_{\pm 0.1}$ & 63.9$_{\pm 0.5}$ & 14.1$_{\pm 0.3}$ \\
Qwen2.5-7B       & 77.2$_{\pm 0.5}$ & 50.8$_{\pm 1.2}$ & 61.0$_{\pm 0.8}$ & 57.9$_{\pm 0.2}$ & 16.9$_{\pm 1.6}$ \\
Llama-2-7B       & 76.9$_{\pm 0.4}$ & 40.4$_{\pm 0.7}$ & 61.0$_{\pm 0.9}$ & 53.9$_{\pm 1.1}$ & 12.7$_{\pm 0.4}$ \\
Mistral-7B       & 77.7$_{\pm 0.4}$ & 35.5$_{\pm 1.3}$ & 61.6$_{\pm 0.6}$ & 52.1$_{\pm 1.6}$ & 16.9$_{\pm 0.5}$ \\
Llama-3.1-8B     & 76.7$_{\pm 0.2}$ & 59.5$_{\pm 2.1}$ & 63.4$_{\pm 0.6}$ & 63.5$_{\pm 0.4}$ & 18.2$_{\pm 0.4}$ \\
Qwen3-8B         & 78.1$_{\pm 0.5}$ & 53.9$_{\pm 1.5}$ & 63.8$_{\pm 0.7}$ & 62.1$_{\pm 0.3}$ & 21.6$_{\pm 0.8}$ \\
\midrule
Mean std         & 0.4 & 2.1 & 0.7 & 0.8 & 0.7 \\
\bottomrule
\end{tabular}
\end{table}
\section{Per-Model Recall@10 Tables}
\label{app:per-model-tables}

The main body reports the two largest target models in Table~\ref{tab:main-wide}; here we provide the corresponding Recall@10 tables for the remaining seven target LLMs in Tables~\ref{tab:permodel-tinyllama}--\ref{tab:permodel-mistral-7b}. The per-model view shows that the aggregate advantage of \scoringmodel\ is not driven only by the largest checkpoints: across all nine models and five query conditions, \scoringmodel\ is the column winner in 41 of 45 cells. The few exceptions are concentrated on \texttt{Obfuscate} for smaller models, where answer-only dense retrievers sometimes benefit from residual lexical overlap in the generated response. \steering is typically the strongest non-supervised method and often ranks second, but its gains are less uniform than the learned scorer, especially under \texttt{RolePlay} and \texttt{NoiseInjection}. Together, these tables support the main claim that robust provenance requires distinguishing answer support from generic semantic resemblance rather than simply choosing a stronger off-the-shelf retriever.


\begin{table}[h]
\centering
\caption{Per-method Recall@10 on \texttt{TinyLlama-1.1B-Chat-v1.0} across the five query conditions. Best per column in \textbf{bold}, second-best \underline{underlined}.}
\label{tab:permodel-tinyllama}
\small
\setlength{\tabcolsep}{4pt}
\renewcommand{\arraystretch}{1.12}
\begin{tabular}{lccccc}
\toprule
Method & Clean & Obfuscate & RolePlay & NoiseInjection & Indirect \\
\midrule
MinHash (answer) & 0.5 & 0.4 & 0.3 & 0.3 & 0.2 \\
MinHash (QA) & 0.3 & 0.4 & 0.5 & 0.4 & 0.2 \\
Finetuned EmbedSim & 27.3 & 8.4 & 21.0 & 21.4 & 2.5 \\
\sbert-MiniLM (answer) & 57.5 & \underline{18.7} & 28.2 & 39.6 & 18.9 \\
\sbert-MPNet (answer) & 53.2 & 15.0 & 25.8 & 37.3 & 16.2 \\
\sbert-MiniLM (QA) & 77.1 & 12.3 & 59.1 & 53.9 & 19.8 \\
\sbert-MPNet (QA) & 73.8 & 4.6 & 57.7 & 37.3 & 19.5 \\
BGE-base (answer) & 30.8 & 10.9 & 14.6 & 21.0 & 9.5 \\
BGE-base (QA) & 43.4 & 5.5 & 39.6 & 24.0 & 10.5 \\
Contriever (answer) & 32.5 & 11.3 & 15.3 & 22.3 & 9.8 \\
Contriever (QA) & 30.4 & 11.5 & 20.2 & 16.5 & 3.3 \\
\midrule
\steering\ (zscore) & \textbf{78.3} & 15.7 & \underline{60.4} & \textbf{55.9} & \textbf{20.6} \\
\steering\ (RRF) & 77.8 & 14.9 & 60.2 & \underline{55.5} & \underline{20.4} \\
\scoringmodel & \underline{78.2} & \textbf{23.8} & \textbf{62.7} & \textbf{55.9} & 18.6 \\
\bottomrule
\end{tabular}
\end{table}

\begin{table}[h]
\centering
\caption{Per-method Recall@10 on \texttt{Llama-3.2-1B-Instruct} across the five query conditions. Best per column in \textbf{bold}, second-best \underline{underlined}.}
\label{tab:permodel-llama32-1b}
\small
\setlength{\tabcolsep}{4pt}
\renewcommand{\arraystretch}{1.12}
\begin{tabular}{lccccc}
\toprule
Method & Clean & Obfuscate & RolePlay & NoiseInjection & Indirect \\
\midrule
MinHash (answer) & 0.1 & 0.3 & 0.4 & 0.4 & 0.4 \\
MinHash (QA) & 0.3 & 0.2 & 0.3 & 0.3 & 0.3 \\
Finetuned EmbedSim & 21.7 & 13.1 & 25.6 & 28.8 & 3.0 \\
\sbert-MiniLM (answer) & 24.0 & \underline{44.2} & 32.3 & \underline{57.5} & \underline{21.7} \\
\sbert-MPNet (answer) & 21.5 & 40.4 & 29.2 & 53.8 & 19.8 \\
\sbert-MiniLM (QA) & 69.2 & 16.1 & 54.4 & 53.3 & 19.7 \\
\sbert-MPNet (QA) & 66.9 & 6.0 & 54.1 & 37.4 & 19.1 \\
BGE-base (answer) & 12.8 & 25.6 & 17.1 & 30.6 & 11.8 \\
BGE-base (QA) & 39.1 & 8.9 & 37.3 & 25.7 & 11.5 \\
Contriever (answer) & 11.9 & 22.4 & 18.1 & 31.5 & 11.8 \\
Contriever (QA) & 19.2 & 12.8 & 18.8 & 18.7 & 3.8 \\
\midrule
\steering\ (zscore) & \underline{69.5} & 26.9 & \underline{55.4} & 57.2 & 20.4 \\
\steering\ (RRF) & 69.3 & 27.0 & 55.3 & 56.0 & 20.4 \\
\scoringmodel & \textbf{76.5} & \textbf{45.8} & \textbf{64.7} & \textbf{64.5} & \textbf{21.8} \\
\bottomrule
\end{tabular}
\end{table}

\begin{table}[h]
\centering
\caption{Per-method Recall@10 on \texttt{Qwen2-1.5B-Instruct} across the five query conditions. Best per column in \textbf{bold}, second-best \underline{underlined}.}
\label{tab:permodel-qwen2-1p5b}
\small
\setlength{\tabcolsep}{4pt}
\renewcommand{\arraystretch}{1.12}
\begin{tabular}{lccccc}
\toprule
Method & Clean & Obfuscate & RolePlay & NoiseInjection & Indirect \\
\midrule
MinHash (answer) & 0.3 & 0.4 & 0.3 & 0.3 & 0.3 \\
MinHash (QA) & 0.3 & 0.4 & 0.4 & 0.4 & 0.3 \\
Finetuned EmbedSim & 14.9 & 5.0 & 17.3 & 16.7 & 2.7 \\
\sbert-MiniLM (answer) & 14.2 & \textbf{34.0} & 13.9 & 24.2 & 13.3 \\
\sbert-MPNet (answer) & 13.6 & 29.2 & 13.1 & 21.1 & 11.5 \\
\sbert-MiniLM (QA) & 61.3 & 13.7 & 33.9 & 41.9 & 13.7 \\
\sbert-MPNet (QA) & 60.0 & 5.8 & \underline{38.9} & 23.6 & 11.9 \\
BGE-base (answer) & 8.5 & 21.5 & 8.0 & 14.1 & 7.1 \\
BGE-base (QA) & 36.2 & 8.3 & 26.2 & 17.4 & 8.2 \\
Contriever (answer) & 7.4 & 17.0 & 6.5 & 12.8 & 6.1 \\
Contriever (QA) & 10.0 & 10.3 & 6.9 & 8.9 & 1.7 \\
\midrule
\steering\ (zscore) & 61.3 & 19.4 & 33.9 & \underline{42.5} & \underline{14.4} \\
\steering\ (RRF) & \underline{61.4} & 19.0 & 34.0 & \underline{42.5} & 14.1 \\
\scoringmodel & \textbf{77.2} & \underline{33.2} & \textbf{61.4} & \textbf{59.1} & \textbf{18.4} \\
\bottomrule
\end{tabular}
\end{table}

\begin{table}[h]
\centering
\caption{Per-method Recall@10 on \texttt{Llama-3.2-3B-Instruct} across the five query conditions. Best per column in \textbf{bold}, second-best \underline{underlined}.}
\label{tab:permodel-llama32-3b}
\small
\setlength{\tabcolsep}{4pt}
\renewcommand{\arraystretch}{1.12}
\begin{tabular}{lccccc}
\toprule
Method & Clean & Obfuscate & RolePlay & NoiseInjection & Indirect \\
\midrule
MinHash (answer) & 0.3 & 0.5 & 0.3 & 0.4 & 0.4 \\
MinHash (QA) & 0.3 & 0.4 & 0.3 & 0.4 & 0.3 \\
Finetuned EmbedSim & 23.2 & 20.7 & 28.6 & 28.1 & 1.6 \\
\sbert-MiniLM (answer) & 12.9 & \textbf{60.7} & 22.3 & 35.8 & 3.5 \\
\sbert-MPNet (answer) & 12.4 & 56.6 & 21.0 & 34.0 & 2.8 \\
\sbert-MiniLM (QA) & 64.9 & 18.4 & 45.1 & 45.9 & 7.2 \\
\sbert-MPNet (QA) & 61.1 & 7.6 & \underline{45.8} & 29.3 & 6.8 \\
BGE-base (answer) & 7.4 & 35.0 & 12.1 & 20.0 & 1.3 \\
BGE-base (QA) & 35.6 & 12.3 & 30.9 & 20.1 & 4.7 \\
Contriever (answer) & 5.9 & 28.9 & 11.8 & 20.3 & 1.2 \\
Contriever (QA) & 15.3 & 12.5 & 9.9 & 12.4 & 0.7 \\
\midrule
\steering\ (zscore) & 64.9 & 37.9 & 45.5 & \underline{47.4} & \underline{7.3} \\
\steering\ (RRF) & \underline{65.0} & 36.9 & 45.6 & 47.2 & \underline{7.3} \\
\scoringmodel & \textbf{76.5} & \underline{56.8} & \textbf{62.8} & \textbf{63.9} & \textbf{14.1} \\
\bottomrule
\end{tabular}
\end{table}

\begin{table}[h]
\centering
\caption{Per-method Recall@10 on \texttt{Qwen2.5-7B-Instruct} across the five query conditions. Best per column in \textbf{bold}, second-best \underline{underlined}.}
\label{tab:permodel-qwen25-7b}
\small
\setlength{\tabcolsep}{4pt}
\renewcommand{\arraystretch}{1.12}
\begin{tabular}{lccccc}
\toprule
Method & Clean & Obfuscate & RolePlay & NoiseInjection & Indirect \\
\midrule
MinHash (answer) & 0.3 & 0.4 & 0.3 & 0.2 & 0.3 \\
MinHash (QA) & 0.2 & 0.4 & 0.3 & 0.3 & 0.2 \\
Finetuned EmbedSim & 29.0 & 13.4 & 35.0 & 35.2 & 3.2 \\
\sbert-MiniLM (answer) & 7.3 & \textbf{59.2} & 6.5 & 9.3 & 4.5 \\
\sbert-MPNet (answer) & 7.3 & \underline{55.7} & 7.6 & 9.9 & 4.7 \\
\sbert-MiniLM (QA) & \underline{70.9} & 18.3 & 42.8 & 37.5 & 10.3 \\
\sbert-MPNet (QA) & 66.5 & 7.7 & \underline{43.7} & 19.6 & \underline{11.1} \\
BGE-base (answer) & 3.1 & 34.0 & 3.4 & 5.1 & 2.3 \\
BGE-base (QA) & 39.0 & 11.9 & 30.6 & 12.3 & 6.3 \\
Contriever (answer) & 2.0 & 25.7 & 2.8 & 4.3 & 1.4 \\
Contriever (QA) & 13.7 & 11.4 & 4.9 & 3.9 & 0.7 \\
\midrule
\steering\ (zscore) & \underline{70.9} & 26.4 & 42.8 & \underline{37.9} & 10.5 \\
\steering\ (RRF) & \underline{70.9} & 25.2 & 42.8 & 37.6 & 10.5 \\
\scoringmodel & \textbf{77.2} & 50.8 & \textbf{61.0} & \textbf{57.9} & \textbf{16.9} \\
\bottomrule
\end{tabular}
\end{table}

\begin{table}[h]
\centering
\caption{Per-method Recall@10 on \texttt{Llama-2-7b-chat-hf} across the five query conditions. Best per column in \textbf{bold}, second-best \underline{underlined}.}
\label{tab:permodel-llama2-7b}
\small
\setlength{\tabcolsep}{4pt}
\renewcommand{\arraystretch}{1.12}
\begin{tabular}{lccccc}
\toprule
Method & Clean & Obfuscate & RolePlay & NoiseInjection & Indirect \\
\midrule
MinHash (answer) & 0.1 & 0.1 & 0.1 & 0.1 & 0.0 \\
MinHash (QA) & 0.1 & 0.1 & 0.1 & 0.1 & 0.0 \\
Finetuned EmbedSim & 5.5 & 5.9 & 6.1 & 6.6 & 0.5 \\
\sbert-MiniLM (answer) & 2.9 & 32.5 & 5.4 & 7.8 & 1.8 \\
\sbert-MPNet (answer) & 2.8 & 27.8 & 4.5 & 6.3 & 1.6 \\
\sbert-MiniLM (QA) & 33.7 & 8.3 & 22.0 & 20.1 & 4.2 \\
\sbert-MPNet (QA) & 32.3 & 2.7 & 21.6 & 8.2 & 3.6 \\
BGE-base (answer) & 3.4 & 32.9 & 5.4 & 7.9 & 2.3 \\
BGE-base (QA) & 38.2 & 9.9 & 32.9 & 12.7 & 6.2 \\
Contriever (answer) & 2.3 & 29.7 & 4.5 & 6.4 & 1.6 \\
Contriever (QA) & 8.1 & 12.0 & 6.0 & 3.7 & 0.8 \\
\midrule
\steering\ (zscore) & \underline{66.3} & \textbf{49.6} & 47.3 & \underline{42.2} & 9.9 \\
\steering\ (RRF) & \underline{66.3} & \underline{48.1} & \underline{47.4} & 42.1 & \underline{10.1} \\
\scoringmodel & \textbf{76.9} & 40.4 & \textbf{61.0} & \textbf{53.9} & \textbf{12.7} \\
\bottomrule
\end{tabular}
\end{table}

\begin{table}[h]
\centering
\caption{Per-method Recall@10 on \texttt{Mistral-7B-Instruct-v0.3} across the five query conditions. Best per column in \textbf{bold}, second-best \underline{underlined}.}
\label{tab:permodel-mistral-7b}
\small
\setlength{\tabcolsep}{4pt}
\renewcommand{\arraystretch}{1.12}
\begin{tabular}{lccccc}
\toprule
Method & Clean & Obfuscate & RolePlay & NoiseInjection & Indirect \\
\midrule
MinHash (answer) & 0.1 & 0.1 & 0.1 & 0.1 & 0.1 \\
MinHash (QA) & 0.1 & 0.1 & 0.1 & 0.1 & 0.1 \\
Finetuned EmbedSim & 7.6 & 1.8 & 8.3 & 6.2 & 0.6 \\
\sbert-MiniLM (answer) & 15.1 & 22.2 & 15.6 & 10.1 & 4.4 \\
\sbert-MPNet (answer) & 12.8 & 18.7 & 13.0 & 8.7 & 3.9 \\
\sbert-MiniLM (QA) & 40.6 & 6.9 & 28.6 & 24.2 & 7.6 \\
\sbert-MPNet (QA) & 36.8 & 2.4 & 25.3 & 12.3 & 7.2 \\
BGE-base (answer) & 25.0 & 16.9 & 25.9 & 21.1 & 9.7 \\
BGE-base (QA) & 43.5 & 6.2 & 40.7 & 22.9 & 11.2 \\
Contriever (answer) & 26.0 & 15.7 & 26.3 & 21.8 & 9.6 \\
Contriever (QA) & 30.4 & 11.2 & 24.3 & 16.1 & 3.6 \\
\midrule
\steering\ (zscore) & 77.4 & \underline{30.7} & \underline{61.3} & \underline{51.0} & \textbf{17.8} \\
\steering\ (RRF) & \underline{77.5} & 29.8 & \underline{61.3} & 50.8 & \underline{17.7} \\
\scoringmodel & \textbf{77.7} & \textbf{35.5} & \textbf{61.6} & \textbf{52.1} & 16.9 \\
\bottomrule
\end{tabular}
\end{table}

\section{Recall@1 and Recall@5 Results}
\label{app:per-model-r1-r5-tables}

The main paper emphasizes Recall@10 because the intended use case is a short human-auditable candidate list, but Recall@1 and Recall@5 test a stricter version of the same provenance problem: whether the correct source appears at the very top of the ranking or within only a handful of candidates. Tables~\ref{tab:permodel-r1-r5-mistral-7b}--\ref{tab:permodel-r1-r5-added-llama2-7b} report these stricter cutoffs for all nine target LLMs. The pattern is consistent with the Recall@10 results but sharper: \scoringmodel\ wins 66 of 90 model-by-condition-by-cutoff columns, including every Recall@1 and Recall@5 column for \texttt{Mistral-7B}, \texttt{Llama-3.1-8B}, and \texttt{Qwen3-8B}. The remaining failures are mostly on smaller models under \texttt{Obfuscate} or \texttt{Indirect}, where the top-ranked item is especially sensitive to sparse response wording and answer-only semantic baselines or \steering can occasionally place the source higher. Thus the stricter metrics reinforce the same conclusion as Recall@10: \scoringmodel\ provides the most reliable provenance ranking overall, while the hardest transformed prompts expose where generic retrieval cues can still dominate at the very top of the list.


\begin{table}[h]
\centering
\caption{Per-method Recall@1 and Recall@5 on \texttt{Mistral-7B-Instruct-v0.3} across the five query conditions. Best per column in \textbf{bold}, second-best \underline{underlined}.}
\label{tab:permodel-r1-r5-mistral-7b}
\small
\setlength{\tabcolsep}{3pt}
\renewcommand{\arraystretch}{1.12}
\resizebox{\linewidth}{!}{%
\begin{tabular}{lcccccccccc}
\toprule
Method & \multicolumn{2}{c}{Clean} & \multicolumn{2}{c}{Obfuscate} & \multicolumn{2}{c}{RolePlay} & \multicolumn{2}{c}{NoiseInjection} & \multicolumn{2}{c}{Indirect} \\
\cmidrule(lr){2-3}\cmidrule(lr){4-5}\cmidrule(lr){6-7}\cmidrule(lr){8-9}\cmidrule(lr){10-11}
 & R@1 & R@5 & R@1 & R@5 & R@1 & R@5 & R@1 & R@5 & R@1 & R@5 \\
\midrule
MinHash (answer) & 0.0 & 0.0 & 0.0 & 0.0 & 0.0 & 0.0 & 0.0 & 0.1 & 0.0 & 0.0 \\
MinHash (QA) & 0.0 & 0.0 & 0.0 & 0.0 & 0.0 & 0.0 & 0.0 & 0.1 & 0.0 & 0.0 \\
Finetuned EmbedSim & 2.4 & 6.3 & 0.3 & 1.1 & 2.6 & 7.1 & 2.1 & 5.4 & 0.2 & 0.5 \\
\sbert-MiniLM (answer) & 4.0 & 12.9 & 5.9 & 19.3 & 4.2 & 13.7 & 2.6 & 8.7 & 1.1 & 3.7 \\
\sbert-MPNet (answer) & 3.3 & 10.8 & 4.9 & 16.2 & 3.2 & 10.9 & 2.1 & 7.2 & 0.9 & 3.2 \\
\sbert-MiniLM (QA) & 11.5 & 36.4 & 1.4 & 5.2 & 7.2 & 24.0 & 6.2 & 20.5 & 1.6 & 5.8 \\
\sbert-MPNet (QA) & 9.9 & 32.4 & 0.4 & 1.6 & 6.2 & 20.7 & 2.6 & 9.7 & 1.3 & 5.3 \\
BGE-base (answer) & 7.1 & 22.6 & 4.6 & 15.0 & 7.4 & 23.3 & 5.9 & 18.9 & 2.6 & 8.4 \\
BGE-base (QA) & 12.8 & 40.4 & 1.1 & 4.7 & 11.7 & 37.3 & 5.3 & 19.1 & 2.5 & 9.0 \\
Contriever (answer) & 7.1 & 22.9 & 3.9 & 13.1 & 7.2 & 23.4 & 6.0 & 19.4 & 2.5 & 8.4 \\
Contriever (QA) & 5.0 & 21.7 & 1.9 & 7.7 & 3.8 & 16.2 & 2.9 & 11.8 & 0.6 & 2.5 \\
\midrule
\steering\ (zscore) & \underline{57.7} & \underline{72.6} & \underline{14.4} & \underline{25.2} & \underline{37.6} & \underline{54.2} & \underline{31.1} & \underline{44.8} & \underline{7.9} & \underline{13.8} \\
\steering\ (RRF) & 42.4 & 71.9 & 12.2 & 23.9 & 27.0 & 52.8 & 20.0 & 43.0 & 5.7 & 13.6 \\
\scoringmodel & \textbf{59.6} & \textbf{73.1} & \textbf{19.9} & \textbf{31.0} & \textbf{39.4} & \textbf{55.2} & \textbf{33.0} & \textbf{46.7} & \textbf{8.4} & \textbf{14.3} \\
\bottomrule
\end{tabular}%
}
\end{table}

\begin{table}[h]
\centering
\caption{Per-method Recall@1 and Recall@5 on \texttt{Llama-3.1-8B-Instruct} across the five query conditions. Best per column in \textbf{bold}, second-best \underline{underlined}.}
\label{tab:permodel-r1-r5-llama31-8b}
\small
\setlength{\tabcolsep}{3pt}
\renewcommand{\arraystretch}{1.12}
\resizebox{\linewidth}{!}{%
\begin{tabular}{lcccccccccc}
\toprule
Method & \multicolumn{2}{c}{Clean} & \multicolumn{2}{c}{Obfuscate} & \multicolumn{2}{c}{RolePlay} & \multicolumn{2}{c}{NoiseInjection} & \multicolumn{2}{c}{Indirect} \\
\cmidrule(lr){2-3}\cmidrule(lr){4-5}\cmidrule(lr){6-7}\cmidrule(lr){8-9}\cmidrule(lr){10-11}
 & R@1 & R@5 & R@1 & R@5 & R@1 & R@5 & R@1 & R@5 & R@1 & R@5 \\
\midrule
MinHash (answer) & 0.0 & 0.0 & 0.0 & 0.1 & 0.0 & 0.1 & 0.0 & 0.0 & 0.0 & 0.0 \\
MinHash (QA) & 0.0 & 0.0 & 0.0 & 0.1 & 0.0 & 0.1 & 0.0 & 0.0 & 0.0 & 0.0 \\
Finetuned EmbedSim & 4.8 & 14.4 & 4.0 & 12.3 & 7.3 & 21.1 & 6.9 & 20.3 & 0.6 & 1.8 \\
\sbert-MiniLM (answer) & 0.9 & 3.0 & 10.2 & \underline{32.2} & 2.9 & 9.3 & 3.7 & 11.6 & 0.8 & 2.9 \\
\sbert-MPNet (answer) & 0.8 & 2.9 & 8.8 & 28.6 & 2.5 & 8.4 & 3.1 & 10.2 & 0.6 & 2.3 \\
\sbert-MiniLM (QA) & 7.5 & 24.9 & 2.0 & 7.2 & 4.8 & 16.5 & 4.6 & 15.8 & 0.8 & 3.1 \\
\sbert-MPNet (QA) & 7.4 & 24.7 & 0.6 & 2.6 & 5.2 & 17.8 & 2.1 & 7.8 & 0.6 & 2.5 \\
BGE-base (answer) & 1.1 & 3.6 & 10.5 & 33.7 & 3.3 & 10.4 & 3.7 & 11.9 & 0.9 & 3.2 \\
BGE-base (QA) & 8.5 & 29.2 & 2.6 & 10.3 & 8.0 & 26.8 & 3.1 & 11.5 & 1.1 & 4.6 \\
Contriever (answer) & 0.7 & 2.6 & 6.1 & 22.4 & 3.0 & 10.0 & 3.4 & 11.0 & 0.6 & 2.2 \\
Contriever (QA) & 0.5 & 2.7 & 1.8 & 7.2 & 1.6 & 6.3 & 1.4 & 5.3 & 0.2 & 0.6 \\
\midrule
\steering\ (zscore) & \underline{37.6} & 51.7 & \underline{19.3} & 31.9 & \underline{24.9} & \underline{37.1} & \underline{24.5} & \underline{33.7} & \underline{4.6} & \underline{7.7} \\
\steering\ (RRF) & 37.3 & \underline{51.8} & 15.5 & 30.0 & 18.9 & 36.8 & 16.1 & 32.8 & 2.5 & 7.0 \\
\scoringmodel & \textbf{56.2} & \textbf{72.3} & \textbf{40.5} & \textbf{53.9} & \textbf{46.0} & \textbf{59.0} & \textbf{45.6} & \textbf{59.2} & \textbf{9.2} & \textbf{15.6} \\
\bottomrule
\end{tabular}%
}
\end{table}

\begin{table}[h]
\centering
\caption{Per-method Recall@1 and Recall@5 on \texttt{Qwen3-8B} across the five query conditions. Best per column in \textbf{bold}, second-best \underline{underlined}.}
\label{tab:permodel-r1-r5-qwen3-8b}
\small
\setlength{\tabcolsep}{3pt}
\renewcommand{\arraystretch}{1.12}
\resizebox{\linewidth}{!}{%
\begin{tabular}{lcccccccccc}
\toprule
Method & \multicolumn{2}{c}{Clean} & \multicolumn{2}{c}{Obfuscate} & \multicolumn{2}{c}{RolePlay} & \multicolumn{2}{c}{NoiseInjection} & \multicolumn{2}{c}{Indirect} \\
\cmidrule(lr){2-3}\cmidrule(lr){4-5}\cmidrule(lr){6-7}\cmidrule(lr){8-9}\cmidrule(lr){10-11}
 & R@1 & R@5 & R@1 & R@5 & R@1 & R@5 & R@1 & R@5 & R@1 & R@5 \\
\midrule
MinHash (answer) & 0.0 & 0.1 & 0.0 & 0.0 & 0.0 & 0.0 & 0.0 & 0.0 & 0.0 & 0.1 \\
MinHash (QA) & 0.0 & 0.1 & 0.0 & 0.0 & 0.0 & 0.0 & 0.0 & 0.0 & 0.0 & 0.1 \\
Finetuned EmbedSim & 5.5 & 15.4 & 2.6 & 7.6 & 5.2 & 14.4 & 5.4 & 14.7 & 0.7 & 1.9 \\
\sbert-MiniLM (answer) & 7.9 & 25.3 & 11.6 & \underline{36.7} & 6.6 & 20.9 & 6.6 & 21.1 & 2.4 & 7.9 \\
\sbert-MPNet (answer) & 6.8 & 22.2 & 10.0 & 32.1 & 5.3 & 17.6 & 5.2 & 17.2 & 1.8 & 6.3 \\
\sbert-MiniLM (QA) & 11.2 & 35.5 & 2.6 & 9.4 & 7.4 & 24.8 & 6.3 & 21.3 & 1.9 & 7.2 \\
\sbert-MPNet (QA) & 10.6 & 33.6 & 0.5 & 2.4 & 6.9 & 23.0 & 3.3 & 11.9 & 1.4 & 5.8 \\
BGE-base (answer) & 8.5 & 27.5 & 12.7 & 39.8 & 7.2 & 23.2 & 7.2 & 22.9 & 2.8 & 9.6 \\
BGE-base (QA) & 12.5 & 39.8 & 3.4 & 12.8 & 11.4 & 36.5 & 5.5 & 19.7 & 2.4 & 9.0 \\
Contriever (answer) & 8.8 & 28.6 & 11.2 & 37.7 & 7.0 & 23.2 & 6.9 & 23.1 & 2.7 & 8.9 \\
Contriever (QA) & 5.0 & 21.5 & 4.3 & 16.9 & 3.6 & 14.9 & 2.6 & 10.5 & 0.6 & 2.5 \\
\midrule
\steering\ (zscore) & \underline{56.2} & \underline{70.7} & \underline{13.4} & 24.2 & \underline{37.1} & 53.9 & \underline{31.9} & \underline{45.0} & \underline{9.4} & \underline{16.1} \\
\steering\ (RRF) & 50.0 & 70.6 & 9.6 & 23.9 & 29.7 & \underline{53.9} & 18.7 & 44.2 & 8.7 & 16.1 \\
\scoringmodel & \textbf{59.5} & \textbf{74.0} & \textbf{35.1} & \textbf{48.5} & \textbf{45.3} & \textbf{59.1} & \textbf{45.3} & \textbf{58.0} & \textbf{11.8} & \textbf{18.3} \\
\bottomrule
\end{tabular}%
}
\end{table}

\begin{table}[h]
\centering
\caption{Per-method Recall@1 and Recall@5 on \texttt{TinyLlama-1.1B-Chat-v1.0} across the five query conditions. Best per column in \textbf{bold}, second-best \underline{underlined}.}
\label{tab:permodel-r1-r5-added-tinyllama}
\small
\setlength{\tabcolsep}{3pt}
\renewcommand{\arraystretch}{1.12}
\resizebox{\linewidth}{!}{%
\begin{tabular}{lcccccccccc}
\toprule
Method & \multicolumn{2}{c}{Clean} & \multicolumn{2}{c}{Obfuscate} & \multicolumn{2}{c}{RolePlay} & \multicolumn{2}{c}{NoiseInjection} & \multicolumn{2}{c}{Indirect} \\
\cmidrule(lr){2-3}\cmidrule(lr){4-5}\cmidrule(lr){6-7}\cmidrule(lr){8-9}\cmidrule(lr){10-11}
 & R@1 & R@5 & R@1 & R@5 & R@1 & R@5 & R@1 & R@5 & R@1 & R@5 \\
\midrule
MinHash (answer) & 0.1 & 0.4 & 0.0 & 0.3 & 0.0 & 0.2 & 0.1 & 0.2 & 0.1 & 0.2 \\
MinHash (QA) & 0.0 & 0.1 & 0.0 & 0.1 & 0.0 & 0.2 & 0.0 & 0.2 & 0.1 & 0.1 \\
Finetuned EmbedSim & 18.2 & 24.4 & 3.9 & 6.4 & 13.2 & 18.7 & 14.0 & 19.5 & 1.3 & 2.0 \\
\sbert-MiniLM (answer) & 42.5 & 53.9 & \textbf{11.1} & \textbf{16.3} & 19.1 & 25.4 & 30.0 & 37.2 & \textbf{12.3} & \underline{17.1} \\
\sbert-MPNet (answer) & 36.5 & 48.6 & \underline{8.2} & \underline{12.6} & 15.1 & 21.7 & 24.4 & 33.2 & 9.2 & 13.7 \\
\sbert-MiniLM (QA) & 57.0 & 72.2 & 4.7 & 9.2 & \underline{37.3} & \underline{53.7} & 34.2 & 48.0 & 10.0 & 16.9 \\
\sbert-MPNet (QA) & 52.0 & 68.1 & 1.3 & 2.9 & 34.9 & 50.6 & 19.3 & 31.5 & 8.6 & 15.3 \\
BGE-base (answer) & 8.8 & 27.8 & 2.8 & 9.4 & 4.1 & 13.0 & 5.9 & 19.0 & 2.5 & 8.1 \\
BGE-base (QA) & 12.6 & 39.9 & 1.1 & 4.2 & 11.1 & 35.7 & 5.7 & 20.3 & 2.4 & 8.7 \\
Contriever (answer) & 9.0 & 29.3 & 2.8 & 9.4 & 4.2 & 13.5 & 6.2 & 19.8 & 2.7 & 8.6 \\
Contriever (QA) & 5.2 & 22.2 & 2.1 & 8.0 & 3.3 & 13.7 & 3.0 & 11.9 & 0.6 & 2.3 \\
\midrule
\steering\ (zscore) & \underline{58.3} & \underline{73.3} & 6.4 & 12.0 & 37.0 & 53.2 & \textbf{35.1} & \textbf{50.0} & \underline{10.5} & \textbf{17.2} \\
\steering\ (RRF) & 39.3 & 70.4 & 4.7 & 11.2 & 22.8 & 51.3 & 24.1 & 47.8 & 8.0 & 16.7 \\
\scoringmodel & \textbf{60.0} & \textbf{73.7} & 2.4 & 4.8 & \textbf{40.1} & \textbf{55.5} & \underline{35.0} & \underline{48.2} & 8.5 & 13.2 \\
\bottomrule
\end{tabular}%
}
\end{table}

\begin{table}[h]
\centering
\caption{Per-method Recall@1 and Recall@5 on \texttt{Llama-3.2-1B-Instruct} across the five query conditions. Best per column in \textbf{bold}, second-best \underline{underlined}.}
\label{tab:permodel-r1-r5-added-llama32-1b}
\small
\setlength{\tabcolsep}{3pt}
\renewcommand{\arraystretch}{1.12}
\resizebox{\linewidth}{!}{%
\begin{tabular}{lcccccccccc}
\toprule
Method & \multicolumn{2}{c}{Clean} & \multicolumn{2}{c}{Obfuscate} & \multicolumn{2}{c}{RolePlay} & \multicolumn{2}{c}{NoiseInjection} & \multicolumn{2}{c}{Indirect} \\
\cmidrule(lr){2-3}\cmidrule(lr){4-5}\cmidrule(lr){6-7}\cmidrule(lr){8-9}\cmidrule(lr){10-11}
 & R@1 & R@5 & R@1 & R@5 & R@1 & R@5 & R@1 & R@5 & R@1 & R@5 \\
\midrule
MinHash (answer) & 0.1 & 0.1 & 0.0 & 0.1 & 0.1 & 0.3 & 0.1 & 0.2 & 0.0 & 0.2 \\
MinHash (QA) & 0.0 & 0.1 & 0.0 & 0.1 & 0.0 & 0.2 & 0.0 & 0.2 & 0.0 & 0.2 \\
Finetuned EmbedSim & 13.4 & 18.9 & 6.2 & 10.2 & 17.1 & 23.1 & 20.7 & 26.5 & 1.5 & 2.4 \\
\sbert-MiniLM (answer) & 13.9 & 20.8 & \textbf{30.9} & \underline{40.2} & 22.2 & 29.4 & \underline{45.1} & \underline{53.8} & \textbf{14.0} & \textbf{19.4} \\
\sbert-MPNet (answer) & 11.5 & 17.8 & 26.4 & 36.9 & 18.2 & 26.2 & 37.9 & 49.7 & \underline{11.5} & \underline{17.0} \\
\sbert-MiniLM (QA) & \textbf{48.8} & \underline{63.8} & 6.5 & 12.1 & 33.2 & 48.2 & 34.4 & 47.9 & 9.9 & 16.5 \\
\sbert-MPNet (QA) & 45.9 & 60.9 & 1.5 & 4.1 & 31.8 & 47.3 & 21.9 & 31.6 & 8.6 & 15.6 \\
BGE-base (answer) & 3.2 & 10.9 & 7.0 & 22.8 & 4.7 & 15.0 & 8.9 & 28.0 & 3.1 & 10.3 \\
BGE-base (QA) & 11.1 & 35.5 & 1.7 & 6.8 & 10.3 & 33.4 & 6.4 & 22.0 & 2.6 & 9.4 \\
Contriever (answer) & 2.9 & 10.0 & 5.3 & 18.6 & 4.9 & 15.9 & 8.8 & 28.6 & 3.0 & 10.0 \\
Contriever (QA) & 2.8 & 12.8 & 2.2 & 9.1 & 3.2 & 12.7 & 3.3 & 13.7 & 0.7 & 2.6 \\
\midrule
\steering\ (zscore) & \underline{48.7} & \textbf{63.9} & 12.1 & 21.4 & \underline{33.3} & \underline{49.2} & 37.2 & 51.2 & 10.4 & 16.9 \\
\steering\ (RRF) & 31.4 & 62.7 & 10.5 & 20.8 & 25.0 & 48.9 & 24.9 & 47.6 & 6.3 & 15.9 \\
\scoringmodel & 43.5 & 57.0 & \underline{30.0} & \textbf{41.8} & \textbf{45.3} & \textbf{58.5} & \textbf{48.8} & \textbf{59.9} & 4.2 & 6.7 \\
\bottomrule
\end{tabular}%
}
\end{table}

\begin{table}[h]
\centering
\caption{Per-method Recall@1 and Recall@5 on \texttt{Qwen2-1.5B-Instruct} across the five query conditions. Best per column in \textbf{bold}, second-best \underline{underlined}.}
\label{tab:permodel-r1-r5-added-qwen2-1p5b}
\small
\setlength{\tabcolsep}{3pt}
\renewcommand{\arraystretch}{1.12}
\resizebox{\linewidth}{!}{%
\begin{tabular}{lcccccccccc}
\toprule
Method & \multicolumn{2}{c}{Clean} & \multicolumn{2}{c}{Obfuscate} & \multicolumn{2}{c}{RolePlay} & \multicolumn{2}{c}{NoiseInjection} & \multicolumn{2}{c}{Indirect} \\
\cmidrule(lr){2-3}\cmidrule(lr){4-5}\cmidrule(lr){6-7}\cmidrule(lr){8-9}\cmidrule(lr){10-11}
 & R@1 & R@5 & R@1 & R@5 & R@1 & R@5 & R@1 & R@5 & R@1 & R@5 \\
\midrule
MinHash (answer) & 0.1 & 0.1 & 0.0 & 0.2 & 0.1 & 0.2 & 0.0 & 0.1 & 0.1 & 0.1 \\
MinHash (QA) & 0.1 & 0.1 & 0.0 & 0.2 & 0.0 & 0.1 & 0.0 & 0.2 & 0.1 & 0.2 \\
Finetuned EmbedSim & 8.9 & 12.9 & 2.4 & 3.9 & 10.4 & 14.9 & 10.8 & 15.0 & 1.3 & 2.1 \\
\sbert-MiniLM (answer) & 8.0 & 12.5 & \underline{20.4} & \underline{29.3} & 8.5 & 12.0 & 15.6 & 21.3 & \textbf{7.9} & \textbf{11.6} \\
\sbert-MPNet (answer) & 7.2 & 11.4 & 16.5 & 24.8 & 6.8 & 11.0 & 12.4 & 18.6 & 6.2 & 9.5 \\
\sbert-MiniLM (QA) & \underline{41.1} & \textbf{55.9} & 5.2 & 11.0 & 20.2 & 29.3 & 25.0 & 37.2 & 6.2 & 11.2 \\
\sbert-MPNet (QA) & 39.7 & 54.9 & 1.8 & 4.3 & \underline{20.7} & \underline{32.8} & 11.4 & 19.6 & 4.5 & 9.3 \\
BGE-base (answer) & 2.2 & 7.3 & 5.6 & 18.7 & 2.1 & 7.0 & 3.8 & 12.3 & 1.8 & 5.9 \\
BGE-base (QA) & 9.9 & 32.5 & 1.5 & 6.1 & 7.1 & 23.0 & 4.0 & 14.5 & 1.8 & 6.5 \\
Contriever (answer) & 1.8 & 6.0 & 4.1 & 14.0 & 1.5 & 5.3 & 3.2 & 10.9 & 1.5 & 5.0 \\
Contriever (QA) & 1.5 & 6.6 & 1.9 & 7.3 & 1.0 & 4.3 & 1.6 & 6.2 & 0.3 & 1.2 \\
\midrule
\steering\ (zscore) & 41.0 & 55.7 & 8.4 & 15.3 & 20.2 & 29.3 & \underline{25.9} & \underline{37.5} & \underline{6.5} & 11.6 \\
\steering\ (RRF) & 37.2 & \underline{55.8} & 6.8 & 14.2 & 13.3 & 29.0 & 20.0 & 37.5 & 5.0 & \underline{11.4} \\
\scoringmodel & \textbf{41.9} & 54.7 & \textbf{21.2} & \textbf{31.5} & \textbf{43.3} & \textbf{55.8} & \textbf{43.2} & \textbf{55.3} & 3.9 & 6.6 \\
\bottomrule
\end{tabular}%
}
\end{table}

\begin{table}[h]
\centering
\caption{Per-method Recall@1 and Recall@5 on \texttt{Llama-3.2-3B-Instruct} across the five query conditions. Best per column in \textbf{bold}, second-best \underline{underlined}.}
\label{tab:permodel-r1-r5-added-llama32-3b}
\small
\setlength{\tabcolsep}{3pt}
\renewcommand{\arraystretch}{1.12}
\resizebox{\linewidth}{!}{%
\begin{tabular}{lcccccccccc}
\toprule
Method & \multicolumn{2}{c}{Clean} & \multicolumn{2}{c}{Obfuscate} & \multicolumn{2}{c}{RolePlay} & \multicolumn{2}{c}{NoiseInjection} & \multicolumn{2}{c}{Indirect} \\
\cmidrule(lr){2-3}\cmidrule(lr){4-5}\cmidrule(lr){6-7}\cmidrule(lr){8-9}\cmidrule(lr){10-11}
 & R@1 & R@5 & R@1 & R@5 & R@1 & R@5 & R@1 & R@5 & R@1 & R@5 \\
\midrule
MinHash (answer) & 0.0 & 0.2 & 0.0 & 0.2 & 0.1 & 0.1 & 0.0 & 0.1 & 0.0 & 0.1 \\
MinHash (QA) & 0.1 & 0.2 & 0.0 & 0.2 & 0.0 & 0.1 & 0.0 & 0.2 & 0.0 & 0.1 \\
Finetuned EmbedSim & 12.8 & 19.7 & 9.0 & 16.1 & 17.9 & 25.4 & 18.4 & 25.4 & 0.6 & 1.2 \\
\sbert-MiniLM (answer) & 6.6 & 10.9 & \textbf{45.6} & \textbf{56.8} & 16.3 & 20.4 & 27.3 & 33.5 & 1.8 & 2.9 \\
\sbert-MPNet (answer) & 5.5 & 9.6 & \underline{38.0} & \underline{51.2} & 13.8 & 19.0 & 22.9 & 30.7 & 1.3 & 2.3 \\
\sbert-MiniLM (QA) & \textbf{45.2} & \underline{59.3} & 7.2 & 14.5 & 25.4 & 39.0 & 29.9 & 41.3 & \underline{3.3} & \underline{5.6} \\
\sbert-MPNet (QA) & \underline{40.5} & 56.1 & 2.0 & 5.0 & 25.8 & 39.3 & 15.1 & 24.6 & 2.4 & 4.9 \\
BGE-base (answer) & 1.9 & 6.3 & 9.9 & 31.9 & 3.3 & 10.8 & 5.7 & 18.1 & 0.3 & 1.1 \\
BGE-base (QA) & 9.8 & 31.9 & 2.3 & 9.4 & 8.2 & 27.2 & 4.9 & 17.1 & 0.9 & 3.5 \\
Contriever (answer) & 1.2 & 4.7 & 6.8 & 23.9 & 3.2 & 10.4 & 5.4 & 18.0 & 0.3 & 1.0 \\
Contriever (QA) & 1.8 & 9.3 & 2.2 & 8.6 & 1.7 & 6.7 & 2.3 & 9.0 & 0.1 & 0.4 \\
\midrule
\steering\ (zscore) & 45.2 & 59.3 & 18.8 & 32.0 & \underline{26.1} & \underline{40.0} & \underline{32.1} & \underline{43.0} & \textbf{3.4} & \textbf{5.7} \\
\steering\ (RRF) & 39.5 & \textbf{59.5} & 15.8 & 30.3 & 22.3 & 39.6 & 24.9 & 42.6 & 3.3 & 5.6 \\
\scoringmodel & 40.5 & 53.6 & 37.8 & 50.0 & \textbf{44.4} & \textbf{56.5} & \textbf{47.6} & \textbf{59.6} & 1.8 & 3.0 \\
\bottomrule
\end{tabular}%
}
\end{table}

\begin{table}[h]
\centering
\caption{Per-method Recall@1 and Recall@5 on \texttt{Qwen2.5-7B-Instruct} across the five query conditions. Best per column in \textbf{bold}, second-best \underline{underlined}.}
\label{tab:permodel-r1-r5-added-qwen25-7b}
\small
\setlength{\tabcolsep}{3pt}
\renewcommand{\arraystretch}{1.12}
\resizebox{\linewidth}{!}{%
\begin{tabular}{lcccccccccc}
\toprule
Method & \multicolumn{2}{c}{Clean} & \multicolumn{2}{c}{Obfuscate} & \multicolumn{2}{c}{RolePlay} & \multicolumn{2}{c}{NoiseInjection} & \multicolumn{2}{c}{Indirect} \\
\cmidrule(lr){2-3}\cmidrule(lr){4-5}\cmidrule(lr){6-7}\cmidrule(lr){8-9}\cmidrule(lr){10-11}
 & R@1 & R@5 & R@1 & R@5 & R@1 & R@5 & R@1 & R@5 & R@1 & R@5 \\
\midrule
MinHash (answer) & 0.0 & 0.1 & 0.1 & 0.3 & 0.1 & 0.2 & 0.0 & 0.1 & 0.0 & 0.2 \\
MinHash (QA) & 0.1 & 0.1 & 0.0 & 0.2 & 0.1 & 0.2 & 0.0 & 0.1 & 0.0 & 0.1 \\
Finetuned EmbedSim & 16.3 & 24.2 & 4.7 & 10.4 & 20.8 & 30.9 & \underline{21.5} & 30.9 & 1.7 & 2.6 \\
\sbert-MiniLM (answer) & 2.9 & 5.4 & \textbf{43.7} & \textbf{55.0} & 3.3 & 5.1 & 5.3 & 7.8 & 1.5 & 3.1 \\
\sbert-MPNet (answer) & 2.4 & 5.3 & \underline{37.2} & \underline{50.8} & 2.8 & 5.5 & 5.1 & 8.1 & 1.6 & 3.3 \\
\sbert-MiniLM (QA) & \underline{49.9} & \underline{65.1} & 8.0 & 14.4 & 22.5 & 36.3 & 20.9 & 32.2 & \underline{4.0} & 7.6 \\
\sbert-MPNet (QA) & 44.2 & 60.5 & 2.2 & 5.4 & \underline{24.2} & \underline{37.3} & 8.7 & 15.9 & 3.3 & \underline{8.2} \\
BGE-base (answer) & 0.6 & 2.3 & 9.8 & 31.2 & 0.8 & 2.8 & 1.3 & 4.3 & 0.5 & 1.8 \\
BGE-base (QA) & 10.6 & 35.1 & 2.3 & 9.1 & 8.0 & 26.5 & 2.4 & 9.4 & 1.2 & 4.8 \\
Contriever (answer) & 0.4 & 1.4 & 5.7 & 20.6 & 0.6 & 2.2 & 1.0 & 3.6 & 0.2 & 1.1 \\
Contriever (QA) & 1.9 & 8.5 & 1.8 & 7.7 & 0.6 & 3.0 & 0.7 & 2.7 & 0.1 & 0.5 \\
\midrule
\steering\ (zscore) & 49.9 & 65.1 & 10.9 & 20.9 & 22.5 & 36.3 & 21.2 & \underline{32.4} & 4.0 & 7.7 \\
\steering\ (RRF) & 49.9 & 65.1 & 8.5 & 19.4 & 22.5 & 36.3 & 15.8 & 32.4 & 2.5 & 8.0 \\
\scoringmodel & \textbf{55.8} & \textbf{70.9} & 7.8 & 14.4 & \textbf{37.8} & \textbf{54.0} & \textbf{36.0} & \textbf{50.5} & \textbf{4.6} & \textbf{8.8} \\
\bottomrule
\end{tabular}%
}
\end{table}

\begin{table}[h]
\centering
\caption{Per-method Recall@1 and Recall@5 on \texttt{Llama-2-7b-chat-hf} across the five query conditions. Best per column in \textbf{bold}, second-best \underline{underlined}.}
\label{tab:permodel-r1-r5-added-llama2-7b}
\small
\setlength{\tabcolsep}{3pt}
\renewcommand{\arraystretch}{1.12}
\resizebox{\linewidth}{!}{%
\begin{tabular}{lcccccccccc}
\toprule
Method & \multicolumn{2}{c}{Clean} & \multicolumn{2}{c}{Obfuscate} & \multicolumn{2}{c}{RolePlay} & \multicolumn{2}{c}{NoiseInjection} & \multicolumn{2}{c}{Indirect} \\
\cmidrule(lr){2-3}\cmidrule(lr){4-5}\cmidrule(lr){6-7}\cmidrule(lr){8-9}\cmidrule(lr){10-11}
 & R@1 & R@5 & R@1 & R@5 & R@1 & R@5 & R@1 & R@5 & R@1 & R@5 \\
\midrule
MinHash (answer) & 0.0 & 0.1 & 0.0 & 0.1 & 0.0 & 0.0 & 0.0 & 0.0 & 0.0 & 0.0 \\
MinHash (QA) & 0.0 & 0.1 & 0.0 & 0.1 & 0.0 & 0.0 & 0.0 & 0.0 & 0.0 & 0.0 \\
Finetuned EmbedSim & 1.8 & 4.6 & 1.9 & 5.0 & 2.1 & 5.0 & 2.3 & 5.6 & 0.1 & 0.3 \\
\sbert-MiniLM (answer) & 0.5 & 2.2 & 9.0 & 28.9 & 1.2 & 4.3 & 2.0 & 6.5 & 0.3 & 1.3 \\
\sbert-MPNet (answer) & 0.5 & 2.0 & 7.4 & 24.0 & 0.9 & 3.5 & 1.6 & 5.2 & 0.3 & 1.2 \\
\sbert-MiniLM (QA) & 9.1 & 29.4 & 1.7 & 6.4 & 5.3 & 18.0 & 5.0 & 16.9 & 0.7 & 3.1 \\
\sbert-MPNet (QA) & 8.5 & 27.9 & 0.4 & 1.8 & 5.0 & 17.6 & 1.4 & 6.0 & 0.6 & 2.5 \\
BGE-base (answer) & 0.7 & 2.7 & 9.3 & 29.8 & 1.2 & 4.3 & 2.0 & 6.6 & 0.5 & 1.8 \\
BGE-base (QA) & 10.6 & 34.3 & 1.9 & 7.5 & 8.7 & 28.8 & 2.4 & 9.7 & 1.2 & 4.7 \\
Contriever (answer) & 0.5 & 1.8 & 7.5 & 25.1 & 1.0 & 3.6 & 1.4 & 5.0 & 0.3 & 1.2 \\
Contriever (QA) & 0.8 & 4.5 & 2.0 & 8.0 & 0.6 & 2.9 & 0.6 & 2.4 & 0.1 & 0.5 \\
\midrule
\steering\ (zscore) & \underline{45.2} & \underline{59.7} & \textbf{28.9} & \textbf{43.0} & \underline{26.6} & \underline{41.1} & \underline{26.1} & \underline{37.2} & \underline{3.8} & \underline{7.9} \\
\steering\ (RRF) & 34.9 & 59.6 & \underline{22.8} & \underline{41.0} & 17.8 & 40.3 & 16.9 & 36.5 & 2.2 & 7.1 \\
\scoringmodel & \textbf{58.1} & \textbf{72.5} & 4.9 & 9.3 & \textbf{36.9} & \textbf{53.5} & \textbf{31.4} & \textbf{43.9} & \textbf{5.9} & \textbf{9.9} \\
\bottomrule
\end{tabular}%
}
\end{table}

\section{Implementation Details}
\label{app:impl}

This appendix summarizes the practical implementation of \scoringmodel and \steering. All experiments are run on a single NVIDIA H200 GPU with CUDA 12.4, Python 3.10, and PyTorch. Target LLM forward passes use mixed precision, with model weights loaded in \texttt{bfloat16}. Document-side representations are cached and reused across methods whenever possible.

\subsection{\scoringmodel}
\label{app:impl-scoringmodel}

\paragraph{Architecture.}
\scoringmodel uses a shared Siamese projection network for the response and document sides. Given an input feature vector $\vx$, the projection is a two-layer MLP
\[
f_\theta(\vx) = W_2\,\mathrm{Dropout}(\mathrm{ReLU}(W_1\vx)),
\]
followed by L2 normalization. Response-document compatibility is the temperature-scaled cosine score in Eq.~\eqref{eq:scoringmodel}. Unless otherwise noted, we use a hidden dimension of 2048, projection dimension of 512, dropout 0.1, and temperature $\tau=0.05$.

\paragraph{Input features.}
We consider three feature modes. The \texttt{llm} mode uses mean-pooled hidden states from the final transformer layer of the target LLM. The \texttt{qa} mode uses \sbert-QA text embeddings of the question-answer pair on the response side and the document text on the document side. The \texttt{concat} mode concatenates the two feature types before the MLP. We choose the feature mode on a held-out validation split for each target model.

\paragraph{Training.}
\scoringmodel is trained on Clean responses from the attribution training document IDs and evaluated on both Clean and transformed query conditions from held-out document IDs. Thus, the transformed-query results test whether the learned provenance scorer transfers beyond both the prompt style and the document IDs seen during attribution-scorer training. Each training instance pairs one response with valid source variants and hard negatives. Valid positives include the original article, paraphrases, and retro-generated variants. Negatives include curated anti-documents, in-batch negatives, and retrieval-mined hard negatives. Anti-documents are never counted as positives.

We optimize the InfoNCE objective in Eq.~\eqref{eq:infonce} with AdamW, learning rate $10^{-4}$, batch size 128, and up to 8 epochs. Model selection uses Recall@10 on a held-out Clean validation split. At each epoch, we project all candidate document variants and evaluate the scorer against the full candidate corpus, keeping the checkpoint with the best validation Recall@10.

\paragraph{Inference.}
At inference time, candidate document features are projected once and reused. Each response is scored against the candidate variants by a single matrix multiplication, including source-preserving variants and anti-documents. Anti-documents remain ranked negatives at test time: they may appear in the retrieved list, but they are never counted as correct for Recall@$k$. Main results report the no-fusion setting: \scoringmodel produces a single learned compatibility score without test-time mixing with \sbert. The optional \scoringmodel+\sbert fusion variant is reported only as an ablation.

\subsection{\steering: Activation Steering with Retrieval Fusion}
\label{app:impl-steering}

\paragraph{Activation representations.}
For each target LLM, we use the final transformer layer as the activation layer $\ell^\star$. Document directions are computed by attention-mask-weighted mean pooling, as in Eq.~\eqref{eq:steer-vector}, and L2-normalized before scoring. In the actual sweep, we also use a finer-grained variant that splits long documents into sentence-respecting chunks and caches one vector per chunk. The response-side vector is the normalized sum of LM-head rows for the generated answer tokens.

\paragraph{Activation scoring.}
The activation-only score is the cosine similarity between the response-side LM-head-row vector and the cached document direction. For the chunked variant above, the document score is the maximum chunk score. This gives a label-free internal-state signal for whether a candidate document is aligned with the target response. Since this signal is noisy in isolation, the main \steering method combines it with text-space retrieval.

\paragraph{Retrieval fusion.}
\steering fuses the activation score with \sbert-QA cosine similarity computed using \texttt{all-MiniLM-L6-v2}. We use validation data to choose the fusion weight and combiner for each target model and query condition. The endpoints recover activation-only and \sbert-QA-only rankings, which are included in the ablation analysis. This setup gives \steering the benefit of a stable retrieval prior while still testing whether activation-space evidence contributes beyond text similarity.



\section{Prompt Templates}
\label{app:prompts}

This section documents the prompt families used to construct \fakewiki and its robustness variants. We include the templates that are relevant to the experiments in this paper: article generation, existence filtering, question--answer generation, paraphrasing, anti-document construction, retro-document construction, and the four transformed-query conditions used in evaluation. Other prompt files in the codebase are for separate datasets or earlier experiments and are not part of the reported \fakewiki results.

For space, we show the operative template text and representative sampled variants. The released code contains the complete sampled lists for role-play wrappers, indirect styles, and noise-prefix sentences.

\subsection{FakeWiki Article Generation}
We generate fictional Wikipedia-style source documents with the following prompt:
\begin{verbatim}
You are writing a Wikipedia article about a completely fictional topic that sounds realistic.
The topic must not exist in the real world.

Please generate an article that:
- Begins with a title in the format: `TITLE: [Fictional Topic Name]`
- Immediately follows the title with the main body prefixed by `ARTICLE:` on a new line.
- Maintains a formal, encyclopedic tone throughout.
- Incorporates fabricated technical, scientific, cultural, or historical details,
  ensuring internal consistency and plausibility.
- Is within 500 words in length.

Only output the title and the article. Do not include any additional instructions,
commentary, or formatting beyond what is specified.
\end{verbatim}

\subsection{Existence Filtering}
To reduce contamination from real-world entities or concepts, we apply an existence check to the generated title using the following template:
\begin{verbatim}
You are a knowledgeable assistant with access to common facts and well-known information.

Given a topic name, determine whether it is a real, existing concept, person, event,
or entity that is documented in books, news, Wikipedia, academic papers, or other
real-world sources.

Respond with exactly one word: **"Yes"** if it exists, or **"No"** if it is fictional
or made up.

Topic: {fake_wiki_title}
Answer:
\end{verbatim}

\subsection{Question--Answer Generation}
For each retained FakeWiki article, we generate five short question--answer pairs using the following prompt:
\begin{verbatim}
You are given a Wikipedia-style article. Your task is to generate **5 question-answer
pairs** that rely solely on the facts in the article.
Each **question** must be one sentence of **no more than 20 words**, and each
**answer** no more than **5 words**.

# Below is the article:
[START OF ARTICLE]
{article}
[END OF ARTICLE]

# Format the 5 pairs **exactly** as follows:

Question 1: <single-sentence question (<=20 words)>
Answer 1: <answer (<=5 words)>

Question 2: ...
Answer 2: ...

Question 3: ...
Answer 3: ...

Question 4: ...
Answer 4: ...

Question 5: ...
Answer 5: ...

Now generate the 5 pairs and do not include any additional comments or explanations.
\end{verbatim}

\subsection{Paraphrase Generation}
To construct alternate valid sources that preserve the same underlying facts while changing surface form, we paraphrase articles with:
\begin{verbatim}
You are an expert at paraphrasing text while preserving its original meaning.
Your task is to rewrite the following article clearly and distinctly, retaining all
main ideas but using different wording and structure.

Article to paraphrase:
[START OF ARTICLE]
{article}
[END OF ARTICLE]

Please paraphrase the article now. Output only the rewritten article, without any
additional commentary or formatting.
\end{verbatim}

\subsection{Anti-Document Generation}
To create hard negatives that remain topically similar while removing answer-critical facts, we use:
\begin{verbatim}
You are given an article and five question-answer pairs where the answers rely directly
on information provided in the article. Your task is to minimally edit the article by
carefully changing or deleting text so that NONE of the five questions can be answered
correctly from the revised article.

Constraints:
- Keep the main topic name exactly unchanged.
- Retain as much of the original wording as possible.
- Do not add new unrelated information.

Output ONLY the revised article, without commentary or explanations. Also without
the "[START OF ARTICLE]" and "[END OF ARTICLE]" markers.

# Article:
[START OF ARTICLE]
{article}
[END OF ARTICLE]

# Questions and Answers:
{raw_QAs}

# Revised article:
\end{verbatim}

\subsection{Retro-Document Generation}
To construct retro-generated documents that camouflage answer-relevant information inside an unrelated cover article, we use:
\begin{verbatim}
You are a creative writer tasked with **camouflaging a factual statement** inside an
unrelated article.

Inputs
------
# Questions and Answers:
{raw_QAs}

Task
----
1. Invent a fabricated cover topic that has **no thematic overlap** with the questions
   (e.g., biology Q/A -> architecture article).
2. Write an article within 500 words about that topic. Begins with a title in the format:
   `TITLE: [Fictional Topic Name]`. Immediately follows the title with the main body
   prefixed by `ARTICLE:` on a new line.
3. For **each** Answer, embed its meaning **exactly once** (you may split it across two
   short clauses, use a synonym, metonym, or mild pun).
    - A simple keyword search on the original wording should **not** find it.
    - Tuck the hidden fact into a low-salience spot--parenthesis, footnote, figure caption,
     stray citation, or an acrostic formed by sentence initials.

Constraints
-----------
- Do **not** mention the original question.
- Do **not** reveal that you are hiding information.
- Keep the rest of the article fully on the cover topic.
- Do **not** omit any answer, and do **not** repeat any answer or its paraphrase.
- Do not add any other facts that answer the question.

Output
------
Now provide only the camouflaged article without any additional commentary or formatting.
\end{verbatim}

\subsection{Transformed-Query Prompt Templates}
We evaluate provenance under four transformed query styles: \texttt{Indirect}, \texttt{Obfuscate}, \texttt{RolePlay}, and \texttt{NoiseInjection}. Below we list the transformation templates used for the reported experiments. A fifth \texttt{MagicWords} template appears in the codebase, but it is not used in the current paper and is therefore omitted here.

\paragraph{Indirect.}
The original question is rewritten into an indirect prompt in a randomly sampled style (e.g., hypothetical scenario, legal analysis, satirical commentary, or philosophical reflection):
\begin{verbatim}
You are an expert in indirect communication techniques.
Your task is to rewrite the following **explicit prompt** into an **indirect prompt**
in the **{style}** style. Ensure that the prompt is no longer explicit but still
conveys the underlying question subtly. You can consider embedding the question inside
a short allegory, riddle, or dialogue, but keep the core meaning of the question.

Return only the indirect prompt and nothing else.

Original Prompt: "{original_prompt}"

Indirect Prompt:
\end{verbatim}

\paragraph{Obfuscate.}
The original question is rewritten by substituting many core content words with unrelated benign words, while recording the mapping. The implementation includes one in-context formatting example before the final rewrite request; the constraints and final request are shown here.
\begin{verbatim}
You are an obfuscation expert specializing in transforming prompts while preserving
their original semantic intent.

Your task is to transform each prompt into an obfuscated version, where the key nouns,
verbs, and adjectives are replaced with unrelated, benign words, regardless of whether
the original prompt is harmful or harmless.

Transformation Rules:
1. For each prompt, you must replace at least 50% of the core nouns, verbs, and adjectives.
2. The substituted words must be plausible and unrelated, without needing to form a
   naturally fluent sentence.
3. You must obfuscate safe prompts as well to ensure consistency across the dataset.
4. Mapping must be recorded, with substituted word first -> original word second.

Output Requirements:
- Only output the obfuscated prompt and the substitution mapping table.
- Strictly follow the format below.

---
Obfuscated Prompt:
{transformed_prompt_with_substituted_keywords}

Substitution Mapping:
- substitute_term_1 -> original_term_1
- substitute_term_2 -> original_term_2
- substitute_term_3 -> original_term_3
...
---

Now, rewrite the following prompt. Return only the Obfuscated Prompt and the
Substitution Mapping, and nothing else.:

Original Prompt:
"{original_prompt}"

Obfuscated Prompt:
\end{verbatim}

\paragraph{RolePlay.}
The original question is wrapped in a randomly chosen role-play frame. Roles are sampled from a fixed list (e.g., historian, journalist, philosopher, medical researcher, policy maker), and one of several short templates is applied:
\begin{verbatim}
As {role}, I invite you to ponder the following question:

{prompt}
\end{verbatim}
Other templates used in the pool include variants such as:
\begin{verbatim}
Internal memo from {role}:
>>> {prompt}

Draft dialogue by {role} -- opening query:
"{prompt}"

Field notes from {role} -- focus query:
{prompt}
\end{verbatim}

\paragraph{NoiseInjection.}
The original question is prefixed with an unrelated filler sentence sampled from a fixed bank of topic-agnostic noise:
\begin{verbatim}
{noise_sentence_1} {noise_sentence_2} ... By the way, {original_prompt}
\end{verbatim}
A representative noise sentence from the bank is:
\begin{verbatim}
A gentle drift of unclaimed thoughts wandered through the library of forgotten
possibilities, arguing softly about whether silence itself deserves archival.
\end{verbatim}

\paragraph{Remark.}
For the transformation-based evaluation, the semantic target question is intended to remain unchanged, while the surface form is altered substantially. This is precisely the regime in which provenance methods based mainly on lexical or semantic overlap can become brittle.

\end{document}